\documentclass[10pt,twocolumn,letterpaper]{article}

\usepackage{cvpr}              % To produce the CAMERA-READY version
\usepackage[dvipsnames]{xcolor}

\usepackage{graphicx}
\usepackage{booktabs}
\usepackage{amssymb}
\usepackage{multirow}
\usepackage{makecell}
\usepackage{float}
\usepackage[accsupp]{axessibility}
\definecolor{cvprblue}{rgb}{0.21,0.49,0.74}
\usepackage[pagebackref,breaklinks,colorlinks,citecolor=cvprblue]{hyperref}
\usepackage{listings}
\usepackage[accsupp]{axessibility} % Improves PDF readability for those with disabilities.

\def\paperID{*****} % *** Enter the Paper ID here
\def\confName{CVPR}
\def\confYear{2024}

\title{NTIRE 2024 Restore Any Image Model (RAIM) in the Wild Challenge}

\author{Jie Liang \and Radu Timofte \and Qiaosi Yi \and Shuaizheng Liu \and Lingchen Sun \and Rongyuan Wu \and Xindong Zhang \and Hui Zeng \and Lei Zhang \and Yibin Huang \and Shuai Liu \and Yongqiang Li \and Chaoyu Feng \and Xiaotao Wang \and Lei Lei \and Yuxiang Chen \and Xiangyu Chen \and Qiubo Chen \and Fengyu Sun \and Mengying Cui \and Jiaxu Chen \and Zhenyu Hu \and Jingyun Liu \and Wenzhuo Ma \and Ce Wang \and Hanyou Zheng \and Wanjie Sun \and Zhenzhong Chen \and Ziwei Luo \and Fredrik K. Gustafsson \and Zheng Zhao \and Jens Sjölund \and Thomas B. Schön \and Xiong Dun \and Pengzhou Ji \and Yujie Xing \and Xuquan Wang \and Zhanshan Wang \and Xinbin Cheng \and Jun Xiao \and \and Chenhang He \and Xiuyuan Wang \and Zhi-Song Liu \and Zimeng Miao \and Zhicun Yin \and Ming Liu \and Wangmeng Zuo \and Shuai Li}

\begin{document}
\maketitle
\begin{abstract}
In this paper, we review the NTIRE 2024 challenge on Restore Any Image Model (RAIM) in the Wild. The RAIM challenge constructed a benchmark for image restoration in the wild, including real-world images with/without reference ground truth in various scenarios from real applications. The participants were required to restore the real-captured images from complex and unknown degradation, where generative perceptual quality and fidelity are desired in the restoration result. The challenge consisted of two tasks. Task one employed real referenced data pairs, where quantitative evaluation is available. Task two used unpaired images, and a comprehensive user study was conducted. The challenge attracted more than 200 registrations, where 39 of them submitted results with more than 400 submissions. Top-ranked methods improved the state-of-the-art restoration performance and obtained unanimous recognition from all 18 judges. The proposed datasets are available at \url{https://drive.google.com/file/d/1DqbxUoiUqkAIkExu3jZAqoElr_nu1IXb/view?usp=sharing} and the homepage of this challenge is at \url{https://codalab.lisn.upsaclay.fr/competitions/17632}.
\end{abstract}    
\section{Introduction}
\label{sec:intro}

Image restoration, aiming at recovering high-quality images from their low-quality counterparts, is one of the most popular low-level vision tasks in the research community. However, there has been a large gap between \textbf{A}cademic research and \textbf{I}ndustrial application for a long time. For example, the image signal processing (ISP) systems on digital cameras always face mixed and complex degradations, yet most methods in academic research are designed and evaluated based on simulated and limited degradation. How to design and train a model that can be generalized to practical applications is a challenging yet highly valuable problem.

The deep learning techniques have significantly advanced the performance of image restoration. Recently, generative adversarial networks show good performance in approximating distributions of real photos in image restoration tasks, while the large-scale pre-trained generative diffusion models have provided powerful priors to further improve the quality of image restoration outputs. 

This challenge aims to provide a platform for industrial and academic participants to test and evaluate their algorithms and models on real-world imaging scenarios, bridging the gap between academic research and practical photography. The objectives of this RAIM challenge are:

\begin{itemize}
    \item Construct a benchmark for image restoration in the wild, including real-world images with/without reference ground-truth in various scenarios and objective/subjective evaluation methods;
    \item Promote the research and development of RAIMs with strong generalization performance to images in the wild.
\end{itemize}

\renewcommand{\thefootnote}{}
\footnotetext{Jie Liang, Radu Timofte, Qiaosi Yi, Shuaizheng Liu, Lingchen Sun, Rongyuan Wu, Xindong Zhang, Hui Zeng and Lei Zhang are the organizers of the NTIRE 2024 challenge, and other authors are the participants.}
\footnotetext{The Appendix lists the authors’ teams and affiliations.}
\footnotetext{NTIRE 2024 website: \url{https://cvlai.net/ntire/2024/}}

This challenge is one of the NTIRE 2024 Workshop associated challenges on: dense and non-homogeneous dehazing~\cite{ntire2024dehazing}, night photography rendering~\cite{ntire2024night}, blind compressed image enhancement~\cite{ntire2024compressed}, shadow removal~\cite{ntire2024shadow}, efficient super resolution~\cite{ntire2024efficientsr}, image super resolution ($\times$4)~\cite{ntire2024srx4}, light field image super-resolution~\cite{ntire2024lightfield}, stereo image super-resolution~\cite{ntire2024stereosr}, HR depth from images of specular and transparent surfaces~\cite{ntire2024depth}, bracketing image restoration and enhancement~\cite{ntire2024bracketing}, portrait quality assessment~\cite{ntire2024QA_portrait}, quality assessment for AI-generated content~\cite{ntire2024QA_AI}, restore any image model (RAIM) in the wild~\cite{ntire2024raim}, RAW image super-resolution~\cite{ntire2024rawsr}, short-form UGC video quality assessment~\cite{ntire2024QA_UGC}, low light enhancement~\cite{ntire2024lowlight}, and RAW burst alignment and ISP challenge.

\section{NTIRE 2024 RAIM Challenge}
\label{sec:challenge}

\subsection{Training Data}
In this challenge, participants can train their models using any data they can collect and any pre-trained models they can reach.

\subsection{Validation and Test Data}
To facilitate the design and development of RAIM by participants, we provide two types of validation and test data: paired data with reference ground truth (R-GT), and unpaired data. All data is available now at \url{https://drive.google.com/file/d/1DqbxUoiUqkAIkExu3jZAqoElr_nu1IXb/view?usp=sharing}.

\subsubsection{Paired Data with R-GT}
\begin{figure*}
  \centering
  \includegraphics[width=1\textwidth]{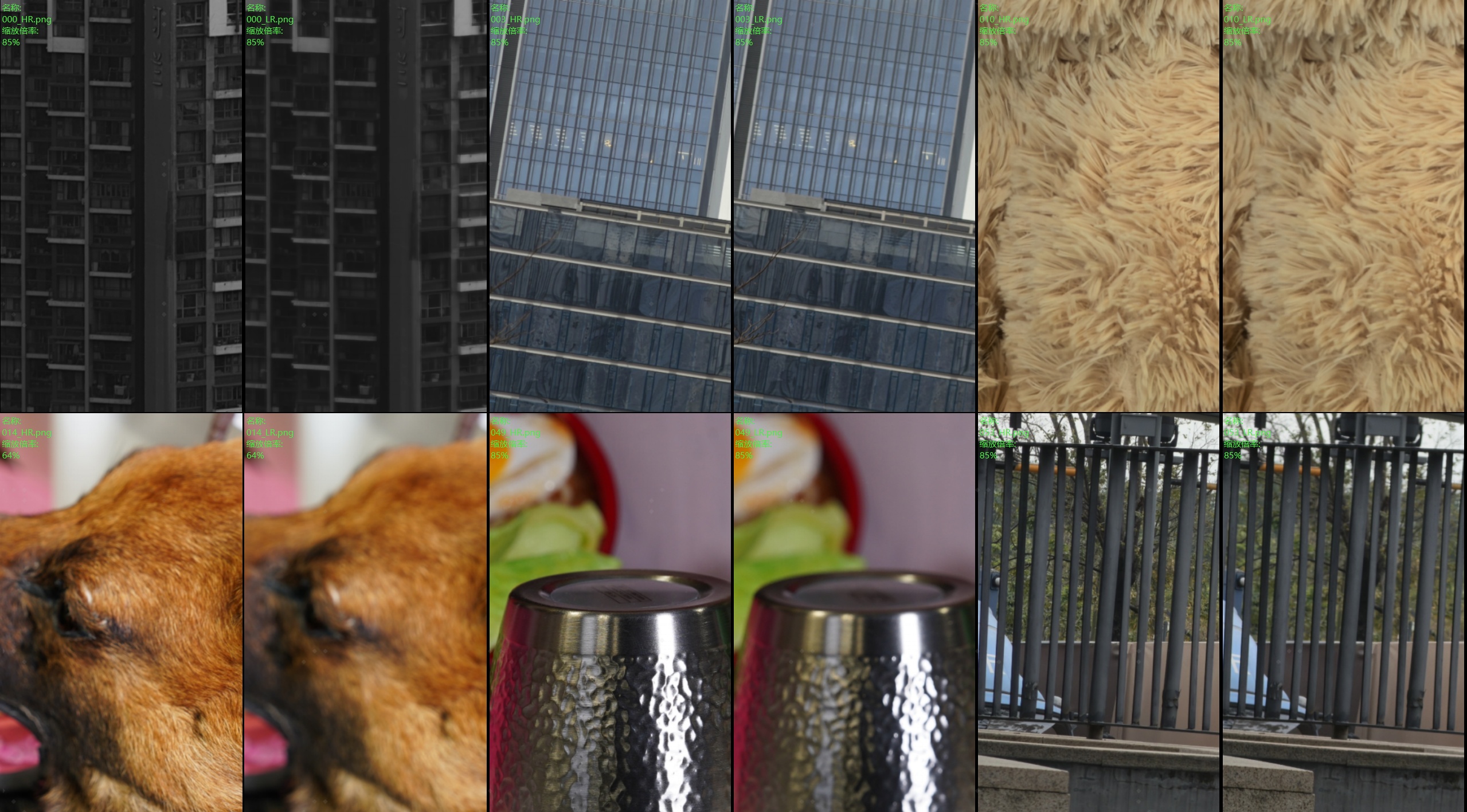} 
  \caption{Example data pairs we have provided.} 
  \label{data1} 
  \vspace{2em}
\end{figure*}

To facilitate the model validation, we first provide some paired data in the following scenarios, where both the input low-quality image and the high-quality R-GT can be collected. Examples can be found in Figure~\ref{data1}.

\noindent\textbf{Image denoising.} Compromising with the size and cost, the photosensitivity of imaging sensors, especially on mobile phone products, is limited. Meanwhile, the illumination of shooting scenes can be poor, especially in low-light imaging. Image denoising is a very fundamental requirement for image restoration.

\noindent\textbf{Image super-resolution.} The focal length of mobile phone cameras is limited, making it hard to meet the needs of continuous and ultra-long magnification zoom. Therefore, mobile cameras will be equipped with digital zoom algorithms, namely super-resolution algorithms.
Out-of-focus restoration. The autofocus (AF) algorithm cannot guarantee 100\% focus accuracy. In fleeting moments of excitement, such as blowing birthday candles, fireworks, etc, the image restoration algorithms are expected to remedy slight defocusing shots.

\noindent\textbf{Motion deblur.} Due to the limitations in aperture size and sensor capability, mobile phone cameras face a trade-off between shutter time and motion blur. A longer shutter may enhance the noise reduction performance, but it is prone to motion blur when encountering foreground object motion or handheld motion. An effective motion deblur algorithm is demanded.

\noindent\textbf{The combinations of the above.} When capturing a real-world photo in the wild, the above issues are usually triggered simultaneously by several factors, such as motion blur in high magnification super-resolution, out-of-focus in low light environments, etc. When multiple problems appear in the image, a strong model is needed to solve them jointly.
In this challenge, we use these data to calculate the full-reference metrics to partially measure the effectiveness of the algorithms and screen the top performers in the early stage.

\subsubsection{Data without R-GT}
\begin{figure*}
  \centering
  \includegraphics[width=1\textwidth]{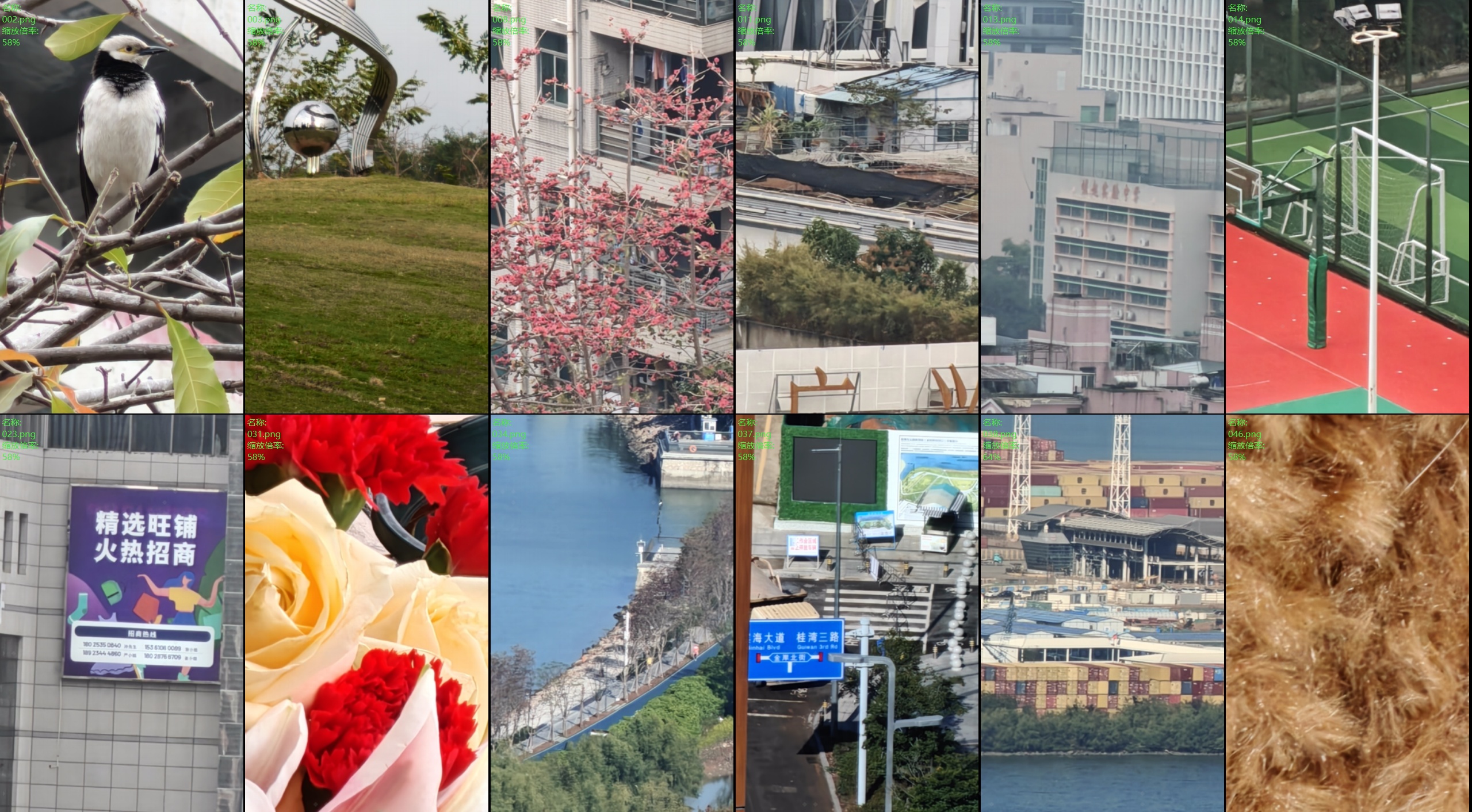} 
  \caption{Example input images we have provided without R-GT.} 
  \label{data2} 
\end{figure*}

In many practical scenarios, the R-GT is very difficult to collect, and the image restoration performance is hard, if not possible, to be measured by full-reference metrics. In this challenge, we also provide the data with the following commonly encountered issues in practice. Examples can be found in Figure~\ref{data2}.

\noindent\textbf{Smoothed details and textures.} Limited by hardware and on-chip computing power, images captured by mobile phone cameras often face a trade-off between noise/artifacts reduction and details/texture richness, impacting the visual quality. However, due to the lack of effective quantitative measures, evaluation can only be done through subjective observation.

\noindent\textbf{Text stroke adhesion in super-resolution.} In the telephoto mode (e.g., equivalent focal length larger than 230mm), shooting small text from a distance is an important yet highly challenging task. Text stroke adhesion, or super-resolution errors (i.e., presenting wrong characters), will greatly deteriorate the user experience.

\noindent\textbf{High light edge and color artifacts.} The optical system of mobile phones is limited, and prone to purple edges, green edges, halos, and fake textures in high-light areas. This problem occurs frequently in reflective scenes, backlight scenes, night scenes, etc., which greatly affects the user's perception.

\noindent\textbf{Low-frequency color noise/blocks/bandings.} The low SNR of the input in mobile phone cameras demands a heavy denoising algorithm to output a clean image. However, due to factors such as computing power and storage, the bit-width of the ISP system is limited. When transitioning from the linear domain to the nonlinear domain, visual color noise, blocks and bandings often appear.

\noindent\textbf{High-frequency aliasing and Moire pattern.} Due to the resolution of imaging sensors, the Moire pattern can appear at specific distances and frequencies. Although users have certain expectations (or understanding) about the appearance of Moire patterns, they still hope to reduce the probability and severity of Moire patterns without affecting the clarity of the image.

\subsection{Evaluation Measures}

We evaluate the effectiveness of the models with both quantitative measures and subjective evaluation.

\subsubsection{Quantitative Measure}
\label{quantitativemeasure}

Following prior arts, we employed the PSNR, SSIM, LPIPS, DISTS and NIQE measures to evaluate the models quantitatively by using the data with R-GT. The evaluation score is computed as follows\footnote{The script of this measure is available at \url{https://drive.google.com/file/d/1Q1CvlbGo-WOgqya5GulS5eYIi2Rgcj5l/view?usp=sharing}.}:

\noindent $SCORE = 20\times \frac{PSNR}{50}+15\times \frac{SSIM-0.5}{0.5}+20\times \frac{1-LPIPS}{0.4}+40\times \frac{1-DISTS}{0.3}+30\times \frac{1-NIQE}{10}$.

\subsubsection{Subjective Evaluation}
\label{qualitativemeasure}

For the test data without R-GT, we judged the perceptual quality of the restored results by visual inspection. Specifically, we invited 18 experienced practitioners and conducted a comprehensive user study. The following features were considered in the evaluation:

\noindent\textbf{Textures and details.} The restored image should have fine and natural textures and details.

\noindent\textbf{Noise.} Noise, especially color noise, should be eliminated. Some luminance noise can be kept to avoid over-smoothness in flat areas.

\noindent\textbf{Artifacts.} Various artifacts, such as worm-like artifacts, color blocks, bandings, over-sharpening, and so on, should be reduced as much as possible.

\noindent\textbf{Fidelity.} The restored image should be loyal to the given input.
More details have been discussed during the competition with all participants by referencing specific images and model outputs.

\subsection{Phases}

\subsubsection{Phase 1: Model Design and Tuning}
In this phase, participants can analyze the given data and tune their models accordingly. We provided:

\begin{itemize}
\item 100 pairs of paired data (i.e., input with R-GT), which can be used to tune the models based on the quantitative measures.

\item 100 images without R-GT, which can be used to tune the model according to visual perception.
\end{itemize}

\subsubsection{Phase 2: Online Feedback}
In this phase, participants can upload their results and get official feedback. We provide:

\begin{itemize}
\item the input low-quality images of another 100 pairs of paired data. 
\end{itemize}

\noindent Only the low-quality input images are provided, and the participants can upload the restoration results to the server and get the quantitative scores online. Users can also upload their results of the images without the R-GT provided in Phase 1 to seek feedback. The organizers will provide feedback to a couple of teams that get the highest quantitative scores of the images with R-GT.

\subsubsection{Phase 3: Final Evaluation}
In this phase, we provide:

\begin{itemize}
\item another 50 images without R-GT for subjective evaluation.
\end{itemize}

\noindent In this phase, we select the top ten teams according to the quantitative score of the 100 images with R-GT in Phase 2, and then arrange a comprehensive user study on their results of the above 50 images without R-GT. The final ranks of the ten teams will be decided based on both the quantitative scores and the subjective user study, with the weight being 40\% and 60\%, respectively.

\subsection{Awards}

The following awards of this challenge are provided:

\begin{itemize}
\item One first-class award (i.e., the champion) with a cash prize of \textbf{US\$1000};

\item Two second-class awards with cash prizes of \textbf{US\$500 each};

\item Three third-class awards with cash prizes of \textbf{US\$200 each}.
\end{itemize}

\subsection{Important Dates}

\begin{itemize}
\item 2024.02.07: Released data of phase 1. Phase 1 began;
\item 2024.02.25: Released data of phase 2. Phase 2 began;
\item 2024.03.17: Released data of phase 3. Phase 3 began;
\item 2024.03.22: Phase 3 results submission deadline;
\item 2024.03.27: Final rank announced.
\end{itemize}
\section{Challenge Results}

In total, the challenge received 200+ registrations, where 39 of them have submitted results in phase 2 with 400+ submissions. In phase 3, we invited the top 12 teams in phase 2 and received 9 valid submissions. Brief illustrations of the methods from participating teams are provided in Section~\ref{teamsandmethods}, while the team information is provided in Section~\ref{appendix}.

\subsection{Phase 2: quantitative comparison on paired data with R-GT}
In phase 2, we got submissions from 30+ teams, where the quantitative results of top-ranked teams are shown in Table~\ref{table_final_result}. The evaluation measure is described in Section \ref{quantitativemeasure}.

\begin{table}\scriptsize
\renewcommand{\arraystretch}{1.2}
\caption{Result of phases 2 and 3, as well as final scores and ranks. We only show teams that participated in phase 3.}
\label{table_final_result}
\centering
\resizebox{\linewidth}{!}{
\begin{tabular}{lcccc}
\toprule
Team&Score in Phase 2&Score in Phase 3&Final Score&Rank\\
\midrule
MiAlgo&79.13&57&91.65&1\\
Xhs-IAG&81.96&47&82.07&2\\
So Elegant&79.69&46&80.09&3\\
IIP\_IR&80.03&14&45.94&4\\
DACLIP-IR&78.65&9&40.03&5\\
TongJi-IPOE&72.99&11&39.91&6\\
ImagePhoneix&78.93&4&34.79&7\\
HIT-IIL&69.80&1&27.92&8\\
\bottomrule
\end{tabular}}
\end{table}

\subsection{Phase 3: qualitative comparison on unpaired data}

In stage three, we invite 18 low-level vision-related students/engineers, who are required to select the top three results of each of the 50 samples. They follow a unified principle as demonstrated in Section~\ref{qualitativemeasure} and the feedback to individual participants. The team information is hidden and the results are randomly shuffled to make fair comparisons. By checking the results of each scorer, we found their opinions are similar so the results are valid. The final score $S_{final}$ is calculated by 
\begin{equation}
S_{final} = 0.4 \times S_2 + 0.6 \times S_3^n,
\end{equation}
where $S_2$ indicates the score in phase 2 and $S_3^n$ denotes the normalized score in phase 3. 

For calculating $S_3^n$, we first calculate the score in phase 3, \ie, $S_3$, where the team is rewarded with 3 points when selected to be top 1, 2 points for the top 2, and 1 point for the top 3. The scores are averaged by 18. Then, we calculate $S_3^n$ by
\begin{equation}
{S_3^n}_{team i} = \frac{S_{3_{team i}} - min(S_3)}{max(S_3) - min(S_3)}.
\end{equation}

\begin{figure*}
  \centering
  \includegraphics[width=1\textwidth]{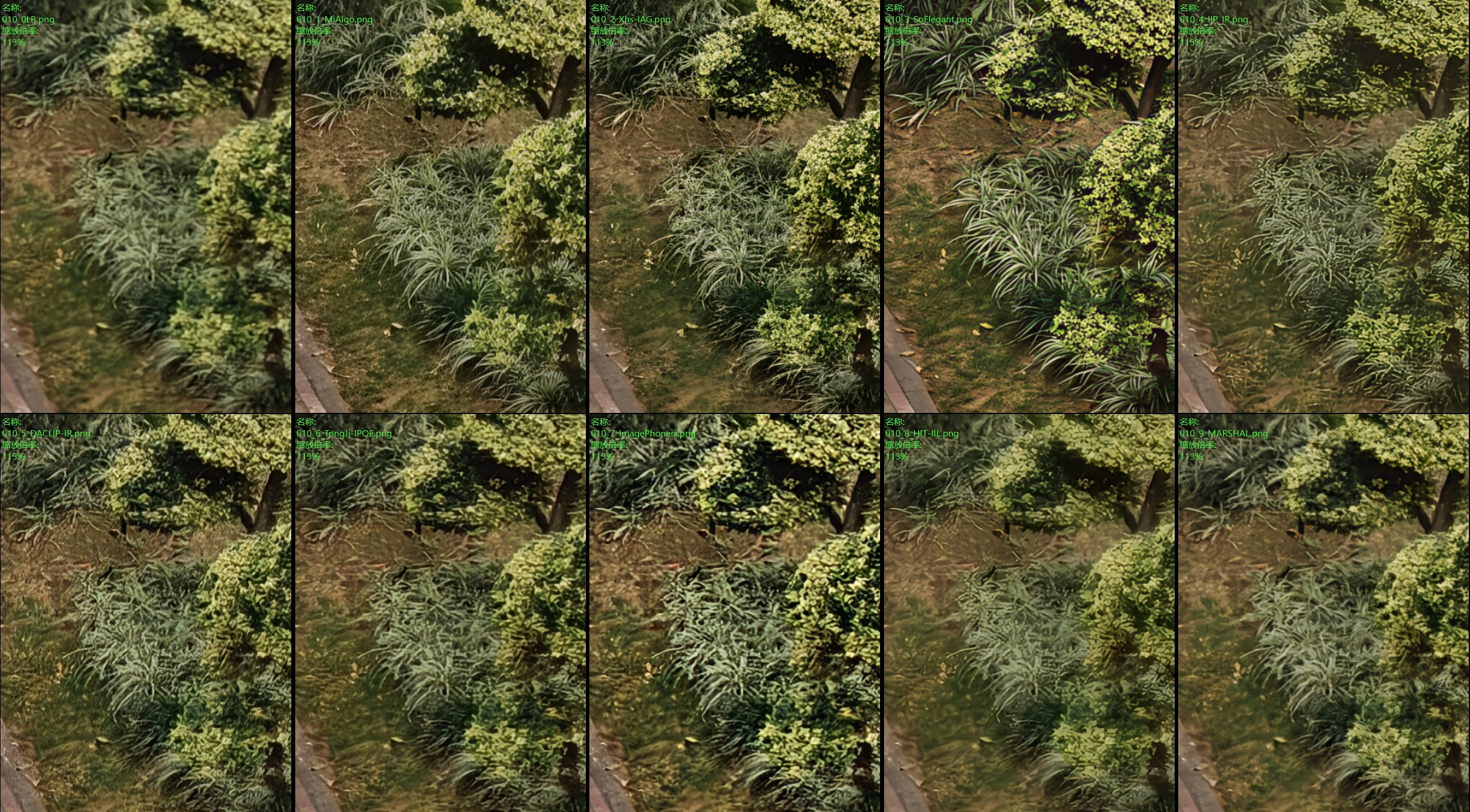} 
  \caption{Visual comparisons of the input LR image (top left) and results from participated teams (others) in phase 3.} 
  \label{result2} 
  \vspace{2em}
\end{figure*}

\begin{figure*}
  \centering
  \includegraphics[width=1\textwidth]{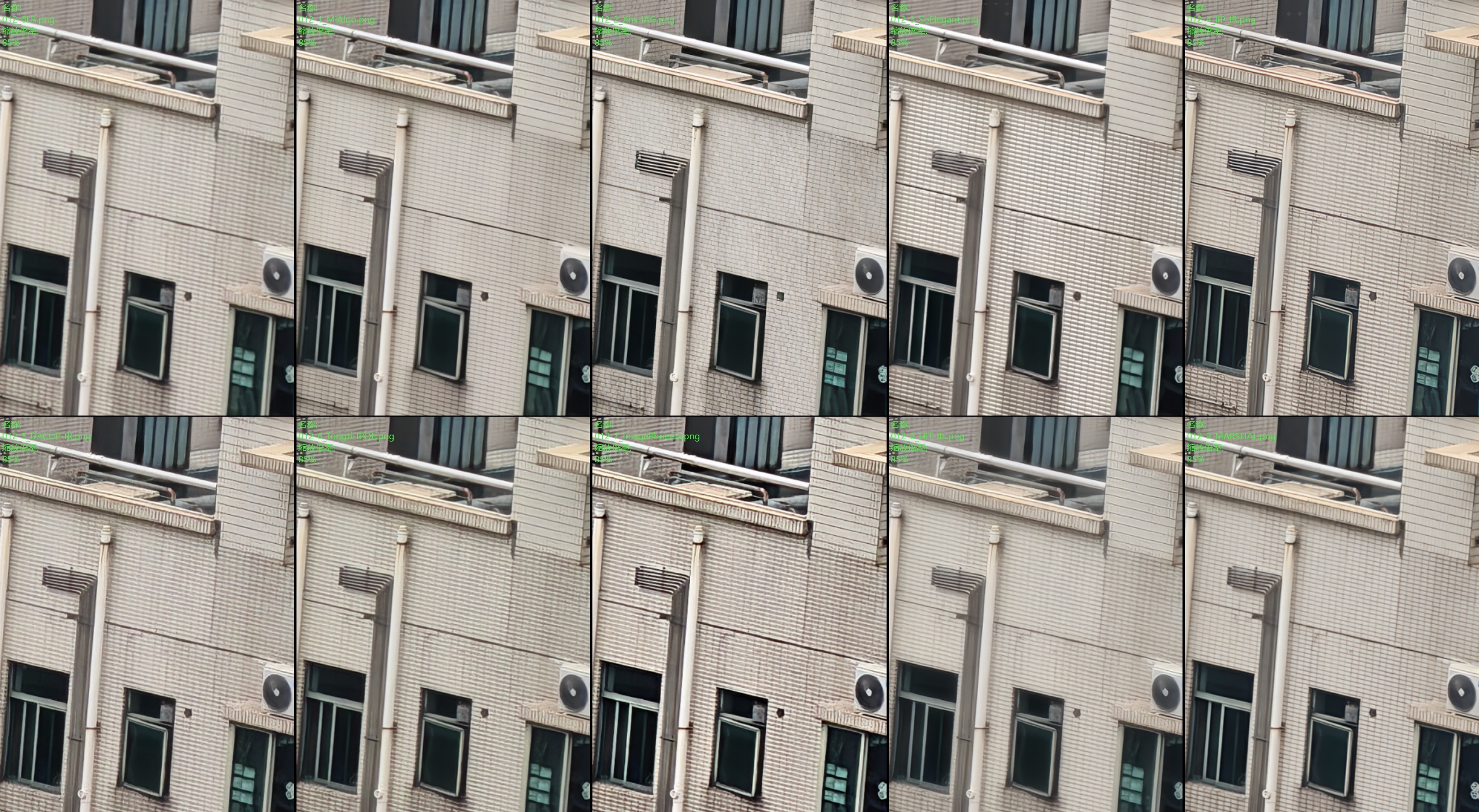} 
  \caption{Visual comparisons of the input LR image (top left) and results from participated teams (others) in phase 3.} 
  \label{result3} 
\end{figure*}

\begin{figure*}
  \centering
  \includegraphics[width=1\textwidth]{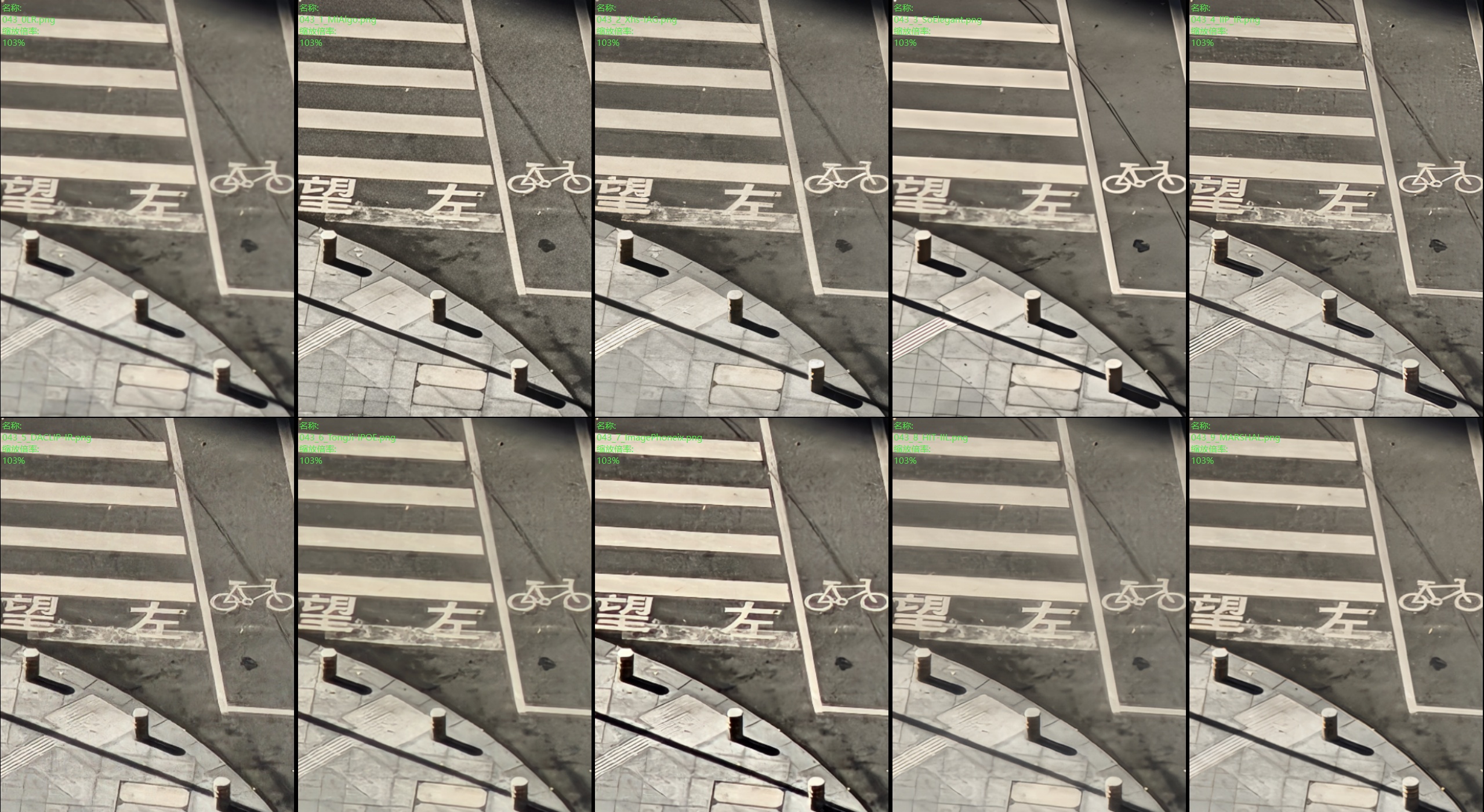} 
  \caption{Visual comparisons of the input LR image (top left) and results from participated teams (others) in phase 3.} 
  \label{result6} 
  \vspace{2em}
\end{figure*}

\begin{figure*}
  \centering
  \includegraphics[width=1\textwidth]{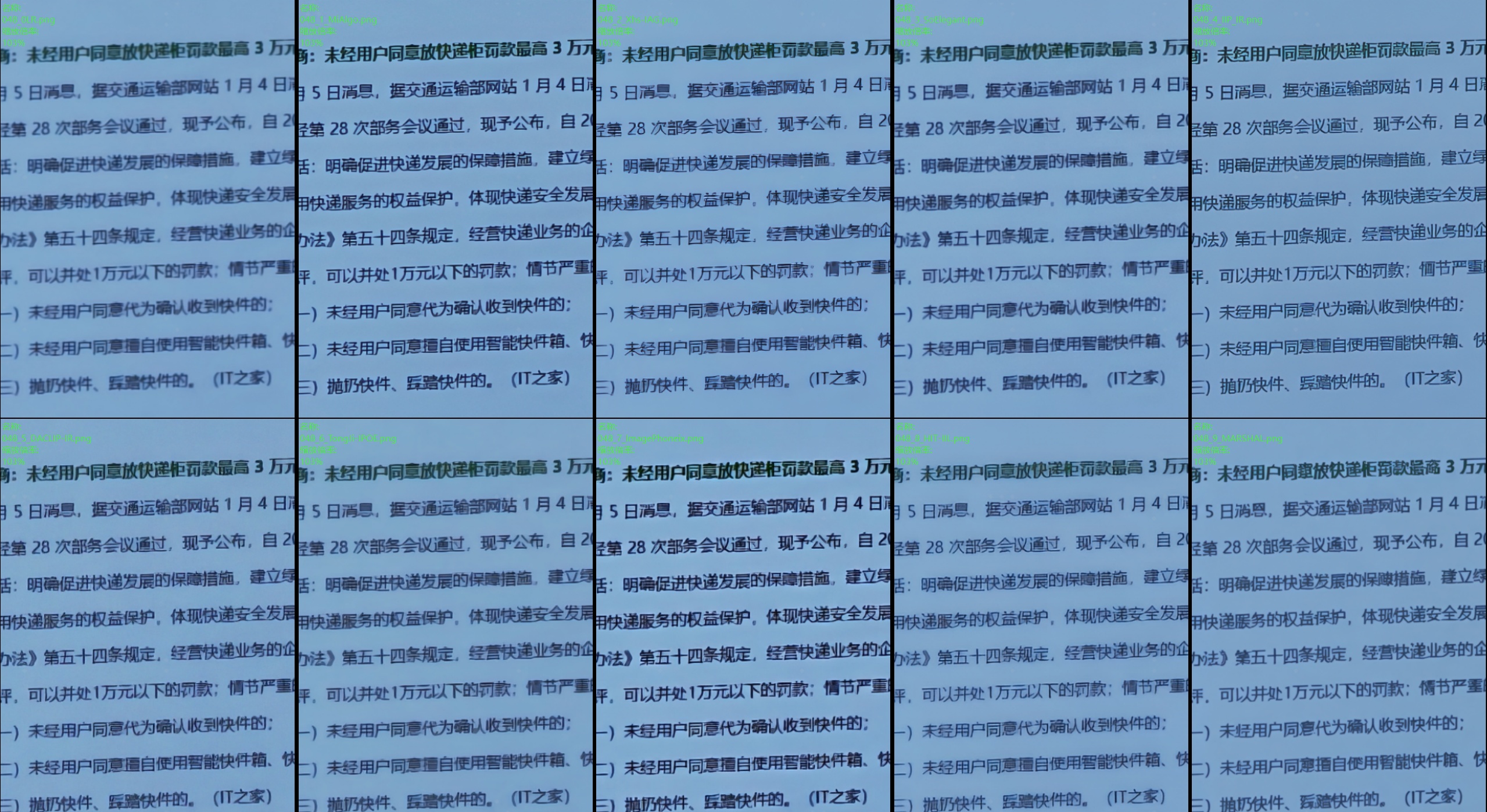} 
  \caption{Visual comparisons of the input LR image (top left) and results from participated teams (others) in phase 3.} 
  \label{result8} 
\end{figure*}

We then show some example visual comparisons in Figures~\ref{result2},~\ref{result3},~\ref{result6} and~\ref{result8}. All visual results in phase 3 are available at \url{https://drive.google.com/file/d/1_vxF2s-WRm59F8Vn1nquE7q4R2zHZTmm/view?usp=sharing}.

% \begin{figure*}
%   \centering
%   \includegraphics[width=1\textwidth]{fig/results/1.png} 
%   \caption{Visual comparisons of the input LR image (top left) and results from participated teams (others) in phase 3.} 
%   \label{result1} 
% \end{figure*}

% \begin{figure*}
%   \centering
%   \includegraphics[width=1\textwidth]{fig/results/4.png} 
%   \caption{Visual comparisons of the input LR image (top left) and results from participated teams (others) in phase 3.} 
%   \label{result4} 
% \end{figure*}

% \begin{figure*}
%   \centering
%   \includegraphics[width=1\textwidth]{fig/results/5.png} 
%   \caption{Visual comparisons of the input LR image (top left) and results from participated teams (others) in phase 3.} 
%   \label{result5} 
% \end{figure*}

% \begin{figure*}
%   \centering
%   \includegraphics[width=1\textwidth]{fig/results/7.png} 
%   \caption{Visual comparisons of the input LR image (top left) and results from participated teams (others) in phase 3.} 
%   \label{result7} 
% \end{figure*}

\section{Teams and Methods}
\label{teamsandmethods}

In this section, we briefly describe the participating teams and their proposed methods.

\subsection{Team MiAlgo}

Team MiAlgo proposed a Wavelet UNet with a Hybrid Transformer and CNN model optimized by adversarial training to tackle the real-world image restoration task.

\subsubsection{Generator model}

\begin{figure*}
  \centering
  \includegraphics[width=1\textwidth]{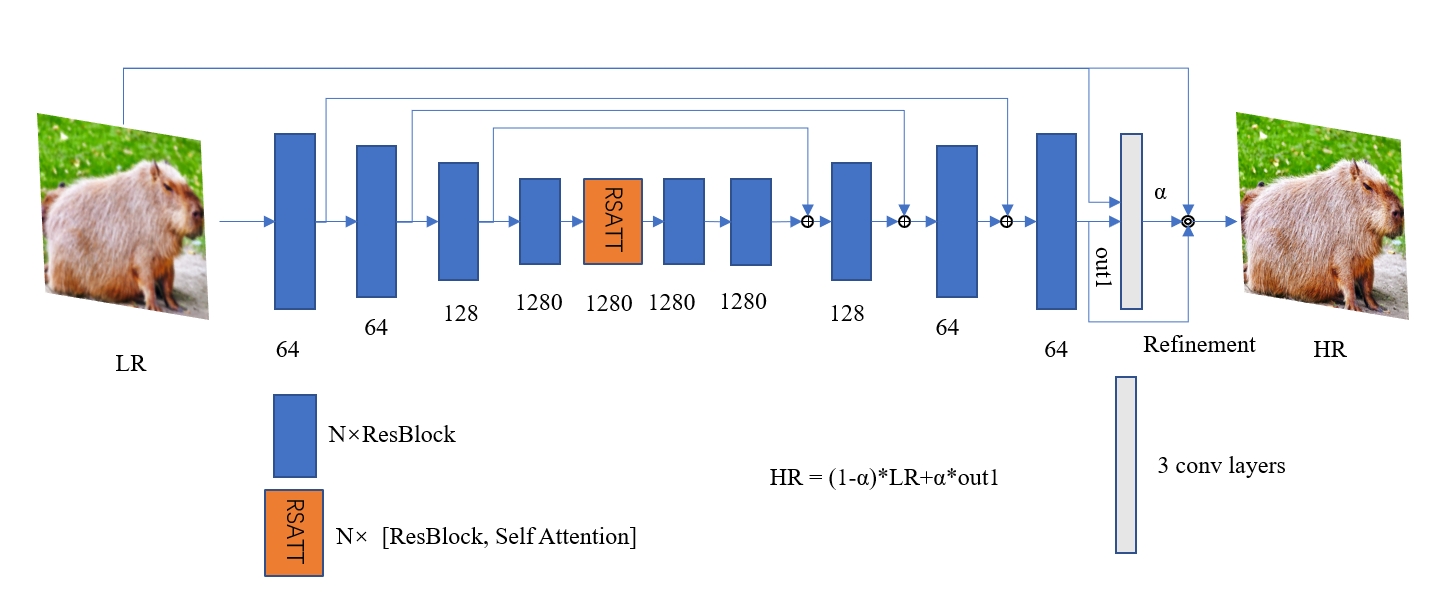} 
  \caption{The overall pipeline of the solution proposed by team MiAlgo.} 
  \label{Figure 1} 
\end{figure*}
 
As shown in Fig. \ref{Figure 1}, the model is based on the MWRCAN \cite{mwcan}. The model uses a UNet architecture that employs Harr wavelet transforms and inverse transforms for 2$\times$ downsampling and upsampling. The major convolution modules consist of $N$ Resblocks, where $N$ is 8 in this case. The channels of the Resblocks are marked in the diagram, and there is also a residual connection in each downsample or upsample block, they omitted these connections for the sake of diagrammatical clarity.

Self-attention in transformers enables the network to identify self-similar features throughout the entire image, thereby enhancing its semantic recognition capabilities. However, the attention structure becomes increasingly time-consuming as the feature size grows, rendering it impractical for high-resolution image restoration tasks. To strike a balance between performance and efficiency, the team integrated RESATT structures into the middle block of UNet. RESATT comprises $N$ basic blocks, each consisting of a res-block followed by a single-head self-attention block.

The UNET produces a 3-channel image called out1. To enhance the quality of the restored image, they incorporate a refinement module based on the EMVD\cite{EMVD} approach. This module helps to recover important details that may have been lost during the restoration process. The refinement module takes in the LR image and out1 as inputs and produces a single-channel fusion weight, denoted by $\alpha$. The final output image is obtained by blending the LR image and out1 using $\alpha$, i.e., $HR=(1-\alpha)LR + \alpha out1$. The refinement module is lightweight, comprising only three convolutional layers with a maximum of 16 channels. Despite its simplicity, it is capable of capturing details that are crucial for the final output image.

The team still insists on using GAN models for general restoration because they have found that diffusion models can lead to unacceptable distortions in text and regular textures. The model has approximately 341MB parameters and takes up 7GB of GPU memory and 180ms to infer a 512$\times$512$\times$3 image on a computer with a 4090GPU. 

\subsubsection{Image degradation}
The official competition only provided 100 pairs of training data, as well as 200 images without ground truth in the validation/test phase. They found that the degradation level of the provided 100 pairs of training data is only consistent with 100 images in phase 2, which is relatively mild. The other images in Phase 2 and Phase 3 have a heavier blurring.

Based on the analysis presented, the team developed two GAN degradation models that introduce varying levels of blurring. They enlarge the generator in \cite{Degradation} by doubling the channels, as the degradation model. The first model was trained with the ESRGAN \cite{esrgan} training method and consisted of 100 pairs of training sets, with high-resolution images serving as input to the degradation GAN model and low-resolution images as ground truth. This model introduced a weak level of blurring.

For the second model, they fine-tuned the weak degradation model using the approach outlined in Ref \cite{Degradation}. They trained this model in an unpaired manner, using 50 high-blurring images from phase 2 as unpaired GT and 1000 high-resolution input images from similar scenes as unpaired input. This model introduced a higher level of blurring compared to the first model. When using the second degradation model, they utilize a human segmentation model and a text segmentation model to segment out the human images with heights <300 pixels and the text with heights <50 pixels. These segments are then replaced with the degradation results from the first degradation model. This strategy helps to reduce the gap between the input and ground truth for small human images and text, and the team has found that this trick improves the fidelity of the results in these regions.

\subsubsection{GAN training} 

The team has an internal ultra-high-definition dataset consisting of approximately 10,000 images. The main scenes include common animals and plants, Chinese and English text, as well as some common urban and rural scenes, which can cover the typical shooting scenarios of mobile phones. They used the two aforementioned degraded GAN models to degrade these images, resulting in a dataset of 20,000 training pairs.

To develop a high-quality image restoration model for phase 2 quantitative measures, they utilized a GAN model trained on 10,000 degraded training pairs from the initial degradation model. The Generator's learning rate was set to 1e-5, with a batch size of 24 and a patch size of 512. The team began training with only L2 loss for $\sim$10,000 iterations, then fixed the loss to include $L2 + 1*PerceptualLoss + 0.1*GANLoss$ for an additional 140,000 iterations. They then fine-tuned the model for $\sim$20,000 iterations with $L2 + 0.1*PerceptualLoss + 0.01*GANLoss +4*LPIPS$ and a lower learning rate of 1e-6 on the official training set (100 pairs) to achieve a slightly higher quantitative score.  The discriminator setting is the same as RealESRGAN \cite{realESRGAN}.

For phase 3, the team continued fine-tuning the model for approximately 100,000 iterations using $loss=L2 + 0.1*PerceptualLoss + 0.01*GANLoss +4*LPIPS$, with a learning rate of $1e-5$. They used a mixed dataset with $80\%$ strong degradation and $20\%$ weak degradation by adjusting the training file list ratio. Finally, they crop each training image into $512\times512$ patches and select the top 10 patches with the higher NIQE score for each image. They continued fine-tuning the model on this subset with a learning rate of $1e-6$ for $\sim$50,000 iterations. Higher NIQE patches generally have richer textures and they found that fine-tuning the model on this subset resulted in better image details.

\subsection{Team Xhs-IAG}

Team Xhs-IAG proposed method by combining SUPIR and DeSRRA, which achieves good generative performance and simultaneously acceptable stability on fidelity.

\subsubsection{Detailed Method Description for Phase2}

\begin{lstlisting}[language={python}]
        window_size = 32,
        embed_dim=180,
        depths=(6, 6, 6, 6, 6, 6),
        num_heads=(6, 6, 6, 6, 6, 6),
        mlp_ratio=4.,
\end{lstlisting}

The dataset they used is LSDIR\cite{li2023lsdir}.
During training, they construct pairs with a resolution of 128x128. The degradation hyperparameters are the same as those for real-esrgan.
They trained 92k iterations with batch size=12 (3 for one GPU, total 4 GPUs) in stage-1 and Adam's learning rate is 1e-4.

In the second stage of training, the team added adversarial loss and perceptual loss, and instead of using lsdir, they \textbf{only} used 100 paired images provided by the official competition. The results show that the degradation distribution of the official evaluation data is close to that of the 100 images.
The specific loss function coefficients are shown below. They trained for a total of 140k iterations in the second stage, with a batch size of 12.  The learning rate for Adam is 5e-5.

\begin{lstlisting}[language={python}]
    discriminator=dict(
        type='UNetDiscriminatorWithSpectralNorm',
        in_channels=3,
        mid_channels=64,
        skip_connection=True),
    pixel_loss=dict(type='L1Loss', loss_weight=1.0, reduction='mean'),
    perceptual_loss=dict(
        type='PerceptualLoss',
        layer_weights={
            '2': 0.1,
            '7': 0.1,
            '16': 1.0,
            '25': 1.0,
            '34': 1.0,
        },
        vgg_type='vgg19',
        perceptual_weight=1.0,
        style_weight=0,
        norm_img=False),
    gan_loss=dict(
        type='GANLoss',
        gan_type='vanilla',
        loss_weight=5e-2,
        real_label_val=1.0,
        fake_label_val=0),
\end{lstlisting}

There is nothing special about the test. For an image, just input it directly into the trained model.

\begin{figure*}[h]
   \centering
    \includegraphics[width=0.8\textwidth]{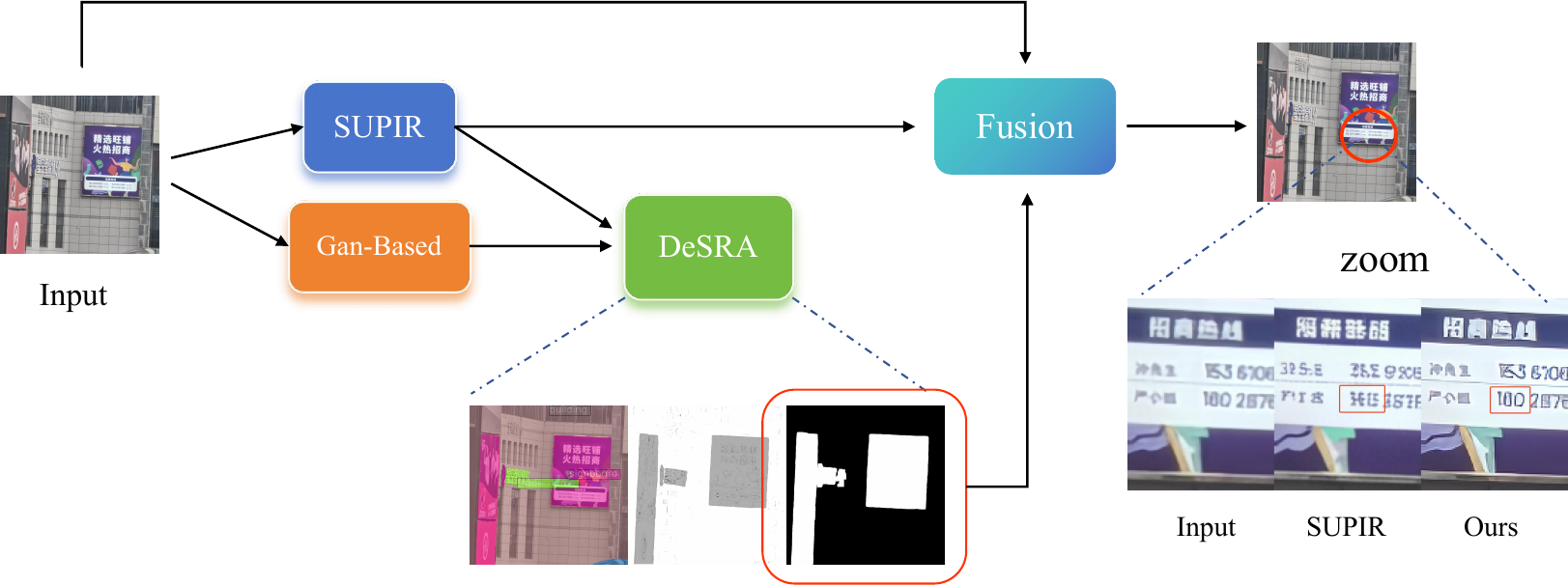}
    \caption{Overall Pipeline of the solution of team Xhs-IAG.}
  \label{fig: pipeline}
\end{figure*}

\subsubsection{Overall Approach}
\label{sec:overall}
In recent years, the diffusion method has achieved remarkable results in the field of image generation, and many methods have recently explored its application in the field of image restoration. Due to the unavailability of data for phase 3 of this competition, the distribution of degradation may differ from phase 2. To increase the generalization ability of our solution, the team use SUPIR\cite{yu2024scaling} as our baseline model.

SUPIR is trained on 20 million images and has good modeling of the distribution of natural images. It supports multiple parameters such as positive prompt, negative prompt, and Classifier free guidance scale to adjust the enhanced results.
Due to the short competition time and the lack of open-source training code for SUPIR, they did not perform any training fine-tuning on SUPIR, but based on its RGB results. To obtain preliminary RGB results, most official default configurations have not been changed. Only the parameters listed in the table \ref{table:Modified} are different from the default parameters.

Although the results generated by diffusion can be natural in most scenes, fidelity issues may arise in some small texture scenes, such as text, patterns, and architectural lines.
Especially in the field of photography, this distortion may be unacceptable to professionals, and even worse than not being processed.
To alleviate this issue, as shown in figure \ref{fig: pipeline}, they will perform another fusion process based on the SUPIR results to obtain the final result. The input of the fusion module includes the SUPIR result, the original image, and a 0/1 mask. To obtain this 0/1 mask, they used the DeSRA\cite{xie2023desra} method. For the sake of fidelity, the fusion module will perform a lighter enhancement on the area with a value of 1 in the mask (e.g., using GAN-based methods), while the area with a value of 0 will be kept as unchanged as possible(i.e., using SUPIR's result).
They introduce our fusion module and DeSRA method in sections \ref{sec:fusion} and \ref{sec:desra} in detail, respectively.

\begin{table}[ht]
\footnotesize
  \caption{Modified config parameters for SUPIR inference }
\label{table:Modified}
  \centering
  \begin{tabular}{lp{2.5cm}p{2.5cm}}
  \toprule
  \textbf{config} &\textbf{default} &\textbf{ours}\\
  \midrule
positive prompt
& Cinematic, High Contrast, highly detailed, taken using a Canon EOS R camera, 
    hyperdetailed photo-realistic maximum detail, 32k, Color Grading, ultra HD, Extreme
     meticulous detailing, skin pore detailing, hyper sharpness, perfect without deformations.
& Cinematic, High Contrast, highly detailed, taken using a Canon EOS R camera, 
             hyperdetailed photo-realistic maximum detail, 32k, Color Grading, ultra HD, extremely meticulous detailing, 
             skin pore detailing, hyper sharpness, perfect without deformations, \textbf{window glass is very clean} 
  \\

\midrule
edm\_steps & 50 & 100 \\
\midrule
sdxl\_ckpt & sd\_xl\_base\_1.0\_0.9vae  & Juggernaut-XL\_v9\_RunDiffusionPhoto\_v2\\
\midrule

s\_cfg & 4.0 & 2.0 \\
  \bottomrule
  \end{tabular}
\end{table}

\subsubsection{Fusion Network}
\label{sec:fusion}
\subsubsection{Architecture}
To ensure the authenticity of the results from diffusion-based models, their fusion module performs fine-tuning based on a binary mask. Specifically, the model takes in three components \textbf{during inference}: the output from SUPIR, the original image, and a binary mask. Areas, where the mask is zero, indicate that the results from SUPIR are already optimal and do not necessitate any modifications, so they will keep this area. Conversely, regions where the mask is one suggest that the results require re-generation to maintain fidelity. They will replace this area with the corresponding LR part to input the model.

\begin{figure}[htb]
  %\begin{center}
%   \vspace{-0.2cm}
   \centering
     % \fbox{\rule{0pt}{2in} \rule{.9\linewidth}{0pt}}
     %\includegraphics[trim={0cm 6.5cm 10cm 0cm},clip,width=1\linewidth]{./img/figure.pdf}
    \includegraphics[width=1.0\linewidth]{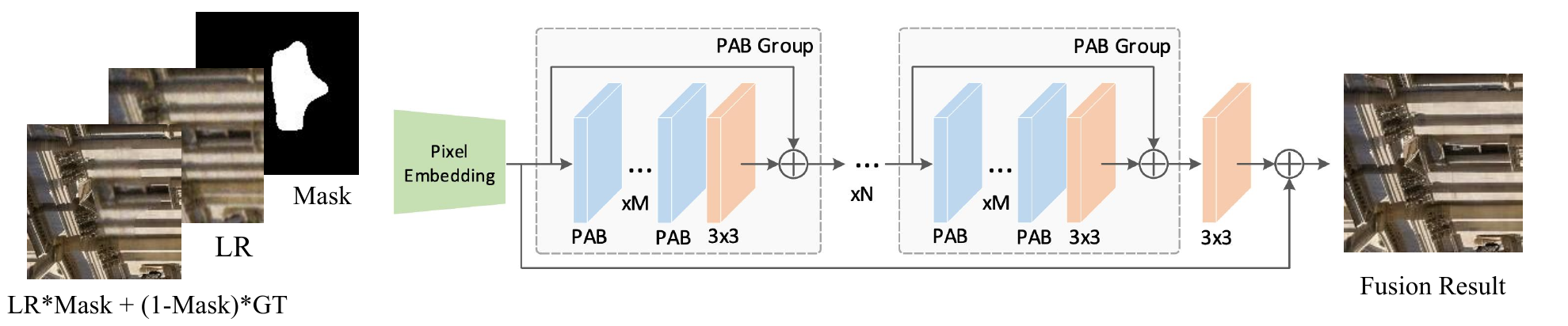}
    \caption{Fusion module of the solution of team Xhs-IAG.}
  \label{fig:fusion}
%   \vspace{-0.4cm}
\end{figure}

In light of the above, the fusion module operates akin to an image inpainting task\cite{zeng2020learning, zeng2022aggregated}, with the key difference that the masked areas are not entirely devoid of information; instead, they contain low-quality images that are awaiting enhancement.
In the training process, the team continue to follow the Real-ESRGAN strategy to generate paired (LR, GT) on the LSDIR dataset.
As illustrated in the figure\ref{fig:fusion}, their model backbone continues to employ SRFormer\cite{zhou2023srformer} (consistent with Phase 2), with the only change being the inputs. At this point, the input will encompass the LR,  mask, as well as the GT and LR combinations derived from the mask. In the inference process, the GT depicted in Figure\ref{fig:fusion} should be substituted with the outcomes yielded by SUPIR.

For the mask used during training, they generate it randomly following the method outlined in STTN\cite{zeng2020learning}, while during testing, they utilize the DeSRA\cite{xie2023desra} approach to obtain the mask. Regarding the DeSRA method, it will be introduced later.

Given that the inputs already contain regions of high quality, the loss function must be correspondingly modified to account for this.
Similar to image inpainting tasks\cite{zeng2022aggregated}, the loss function encompasses hole loss, valid loss, perceptual loss, and adversarial loss. Notably, for the generated fake image, the discriminator employs the technique of using soft labels when calculating the least square loss\cite{zeng2022aggregated}, rather than the hard labels of 0 and 1. This design allows the discriminator to better discern potential mask areas.

\subsubsection{Training Details}
The team used SrFormer\cite{zhou2023srformer} as the backbone, the specific parameters are shown in the following code.
\begin{lstlisting}[language={python}]
    window_size = 24,
    embed_dim = 360,
    depths=(6, 6, 6, 6, 6), 
    num_heads=(6, 6, 6, 6, 6), 
    mlp_ratio=3
\end{lstlisting}

The dataset they used is LSDIR\cite{li2023lsdir}.
During training, the team constructed pairs with a resolution of 144x144. The degradation hyperparameters are the same as those for real-esrgan.
They trained 172k iterations with batch size=8 (2 for one GPU, total 4 GPUs) and Adam's learning rate is 1e-4.
The specific loss function coefficients are shown below.  

\begin{lstlisting}[language={python}]
    valid_loss = dict(type='Valid_loss', loss_weight=0.3),
    hole_loss = dict(type='Hole_loss', loss_weight=0.01),
    perceptual_loss=dict(
        type='PerceptualLoss',
        vgg_type='vgg19',
        layer_weights={
            '1': 1.,
            '6': 1.,
            '11': 1.,
            '20': 1.,
            '29': 1.,
        },
        layer_weights_style={
            '8': 1.,
            '17': 1.,
            '26': 1.,
            '31': 1.,
        },
        perceptual_weight=0.2,
        style_weight=150,
        norm_img=False,
    ),
    gan_loss=dict(
        type='GANLoss',
        gan_type='lsgan',
        loss_weight=0.02,
        real_label_val=1.0,
        fake_label_val=0)
\end{lstlisting}

The generation of random masks during training can be referenced at the specified line in the following GitHub repository: \href{https://github.com/researchmm/STTN/blob/master/core/utils.py#L125}{STTN GitHub Repository}.
        
\subsubsection{DeSRA Method}
\label{sec:desra}
With the fusion model in place, it is necessary to ascertain the masks used during testing, specifically identifying the regions where the diffusion results are distorted.
A straightforward method involves manual annotation of masks, but this approach is not only unfair in the context of competition but also labor-intensive.

The team employ the methodology from DeSRA\cite{xie2023desra} for identifying GAN artifacts, utilizing a combination of structural similarity metrics and semantic segmentation outcomes to generate masks. 
To be precise, they ascertain the mask by contrasting the outputs from the GAN model with those from the diffusion model. The GAN model utilized in this process is the one that has been adequately trained during Phase 2. This choice is motivated by the fact that, despite the GAN model's potential shortcomings in visual quality, it excels in preserving fidelity in intricate details such as text and textures.
By adjusting the parameters, the team strives to align the distribution of the masks with human visual perception. It is important to note that no special parameters are used for any individual image; the same set of parameters is applied consistently across all 50 images.

To enhance the accuracy of the segmentation, they have utilized the Mask2Former model\cite{cheng2022masked} for this task. Compared to the SegFormer model\cite{xie2021segformer} used in the original DeSRA, Mask2Former represents a more advanced approach.
Within the provided code, they have included scripts for mask generation, which encompass all the parameters used, including the weights for semantic categories, contrast\_threshold, area\_threshold, and so on.

\subsection{Team So Elegant}

The team proposed a Consistency-guided Stable Diffusion method for Image Restoration.

\begin{figure}[t]
	\centering
	\includegraphics[width=0.5\textwidth]{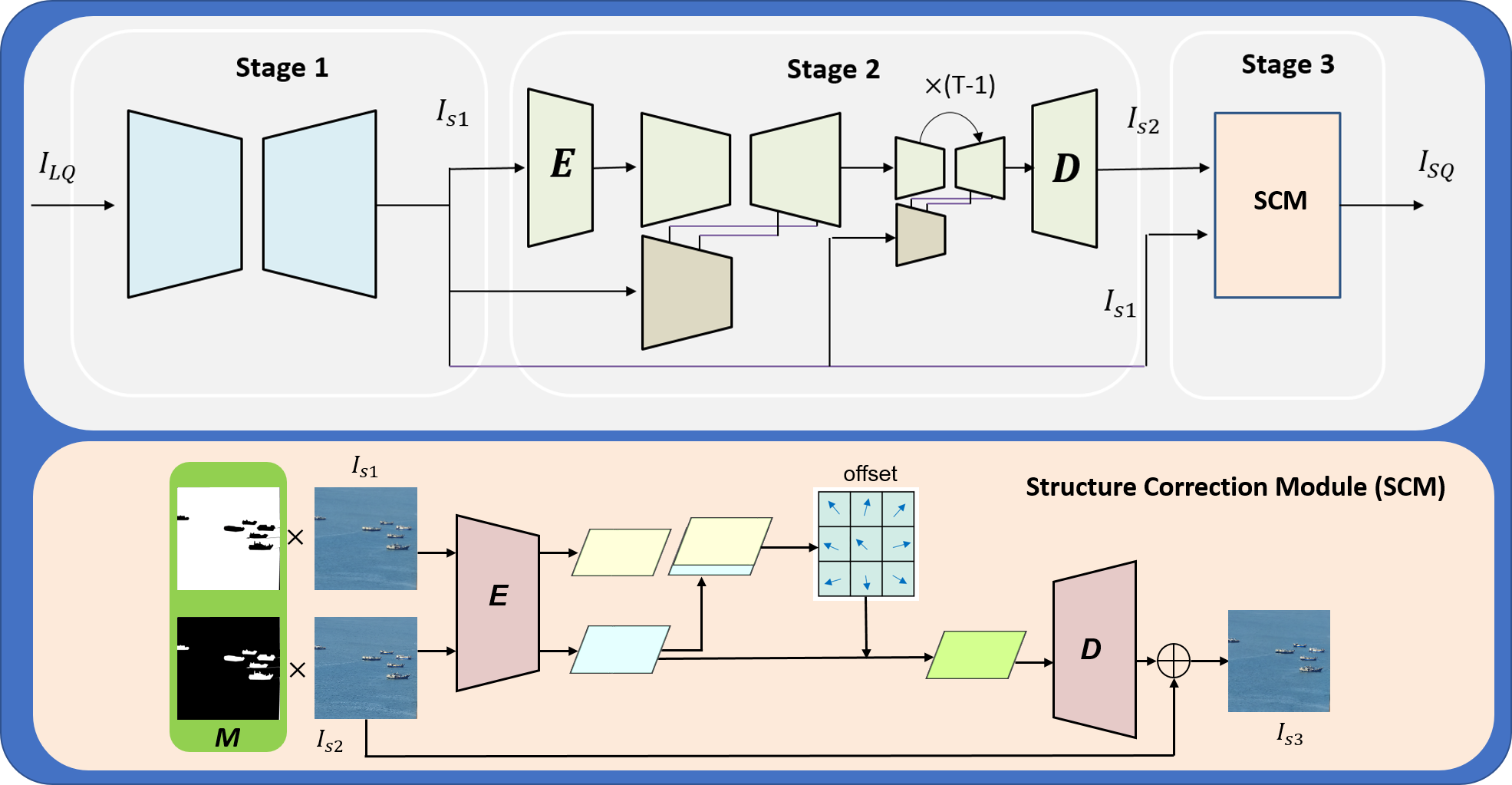}
	\caption{The three-stage pipeline of CGSD proposed by team So Elegant.}
	\label{fig1}
\end{figure}

 As shown in Figure \ref{fig1}, the proposed Consistency Guided Stable Diffusion (CGSD) model has three primary stages. Stage 1 is based on a CNN-based restoration model DiffIR \cite{xia2023diffir} to remove diversified degradations. DiffIR uses the powerful mapping capability of the diffusion model to estimate a compact IR prior representation (IPR) to guide image restoration, thereby improving the recovery efficiency and stability of the diffusion model in image restoration. To bridge the domain gap, the degradation of the given data is used to customize the degradation distribution for training \cite{zhang2023crafting}, which improves the performance of the target test images while maintaining generalization performance. Additionally,  BSRGAN \cite{zhang2021designing} is used to simulate image degradation to generate pairs of data for training. And, virtual focus blur is added to BSRGAN to better suit the target test images. For stage 2, Stable Diffusion (SD) \cite{Rombach_2022_CVPR} is leveraged to refine the texture and details.  To improve the fidelity of SD model restoration, a Consistency-Guided Sampling Module (CGS) is proposed to limit the generation. Specifically, the CGS module takes the recovered image of stage 1 as the consistent  guidance in each decoding step and aligns the recovery results of each step with it:
 \begin{equation}
 	x_{t-1} \gets x_{t-1} + \sigma_{t}(x_{s1}-x_{t-1})
 \end{equation}
where $x_{t-1}$ and $x_{s1}$ corresponds to the noise-free predicted output at step $t-1$ and the recovered $I_{s1}$ latent. $\sigma_t$ represents the weight of the guidance. The image structure is determined in the early diffusion step, and the later stage mainly generates high-frequency details. The final stage 3 is proposed to address the contexture distortion caused by the diffusion model. The contextual information from $I_{s1}$ guides the refined image $I_{s2}$. Similar to \cite{li2020blind}, deformable convolution\cite{dai2017deformable} is employed to warp the details in $I_{s2}$ to match the fidelity of $I_{s1}$. A problematic mask \cite{xie2023desra} $M$ located by a relative local variance distance from $I_{s1}$ and $I_{s2}$ and semantic-aware thresholds are used as the additional condition. The method is implemented in Pytorch and trained using 8x Nvidia V100 GPU for training. For stage 1, the team first uses the original configuration from DiffIR for training and then adjusts the learning rate to 5e-5, batch size to 2, and trains 10K iterations at a resolution of 512x512. For stage 2, they train the SD model using the AdamW \cite{loshchilov2018fixing} optimizer with a learning rate of 1e-4 and a batch size of 64 for 50K steps. For stage 3, they use a batch size of 2 and a patch size of 1024x1024 for training. Adam is used as the optimizer with a learning rate of 1e-4. And they train the model for 20K iterations.

\subsection{Team IIP\_IR}

\begin{figure}
    \centering
    \begin{subcaption}
        \centering
        \includegraphics[width=0.5\textwidth]{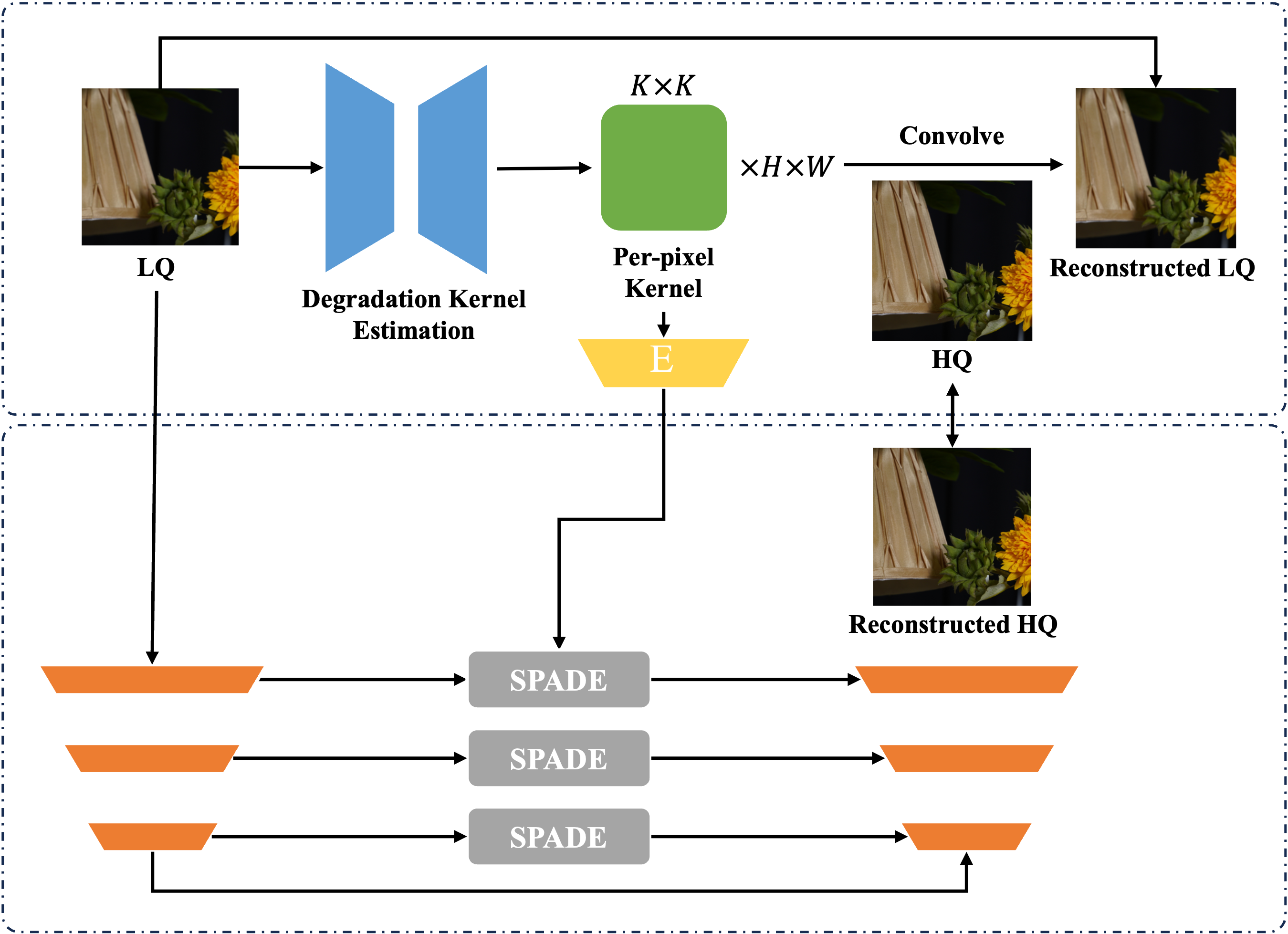}
        \label{fig:phase2}
    \end{subcaption}
    \vfill
    \begin{subcaption}
        \centering
        \includegraphics[width=0.5\textwidth]{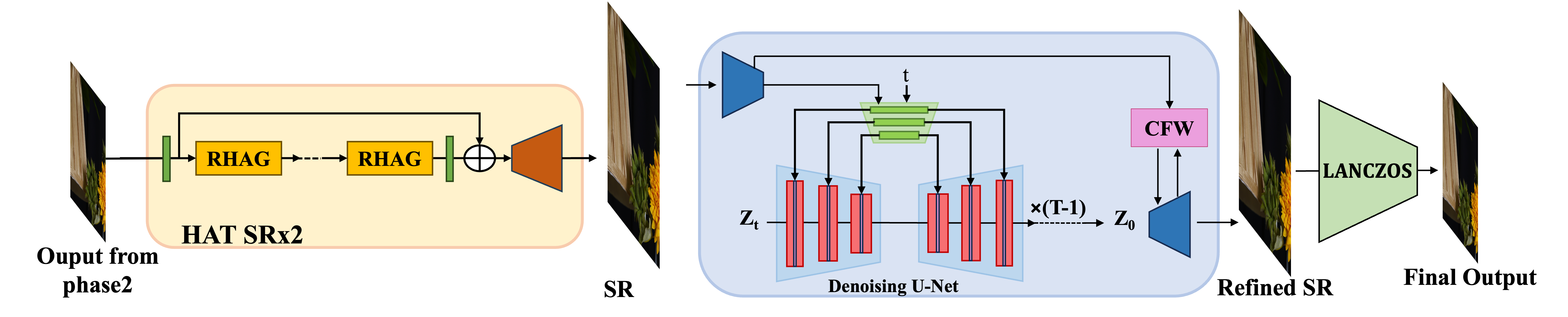}
        \label{fig:phase3}
    \end{subcaption}
    \caption{A visual representation of the solution proposed by IIP\_IR.}
    \label{fig:mainfig}
\end{figure}

The team IIP\_IR has introduced an integrated framework called Degradation-Aware Image Restoration(DAIR) based on the FFTformer architecture introduced in \cite{kong2023efficient} for phase 2. DAIR comprises three main components: Degradation Kernel Estimation (DKE), Degradation Representation Injection (DRI), and FFTfromer. The team’s innovative approach, as illustrated in Figure \ref{fig:phase2}, has the potential to enhance existing models and improve overall performance.

To enable the model to process the degradation of the images, they utilize a method to learn per-pixel degradation convolution kernels similar to blur kernels, which can reconstruct LQ when convolved with HQ images. Unlike the blur kernel, DKE does not constrain the reconstructed kernel to have positive weights that sum to one, thus learning richer degenerate representation.

To maximize the retention of degraded information for image restoration models, the kernels estimated by DKE will be embedded into the Spatially Adaptive representation and injected into U-Net architecture, which is processed through a SPADE module \cite{park2019semantic}. The processing of the SPADE module does not change the network structure, thus DKE and DRI can be applied directly to any Unet-based image restoration model.

In the training process, the team uses the method mentioned in \cite{carbajal2021non} to generate paired data for pre-training the model, improves its generalization ability and adaptability, and finally fine-tunes the model using 100 pairs. While L1 loss normally trained networks which usually produce smooth/blurry results, they apply perception loss and GAN loss constraints to reconstructed LQ and HQ for both the pre-training phase and fine-tuning phase to increase the realism of the image.

Figure \ref{fig:phase3} illustrates the pipeline of phase 3. The team utilizes the model to refine the details of the pre-processed images from phase 2. The images first undergo x2 upscaling using HAT\cite{chen2023activating} to enrich the textures. The initial upscaling phase effectively mitigates distortions of small-scale details such as texts during the texture generation process leveraging pre-trained diffusion priors. They employ StableSR\cite{wang2023exploiting} with SD-Turbo, to further refine the upscaled images, producing realistic textures in regions with severe degradations. The refined images were then downscaled with LANCZOS interpolation to obtain the final output.

\subsection{Team DACLIP-IR}

Team DACLIP-IR proposed a photo-realistic image restoration method with enriched vision-language features.

\begin{figure}[t]
\begin{center}
\includegraphics[width=1.\linewidth]{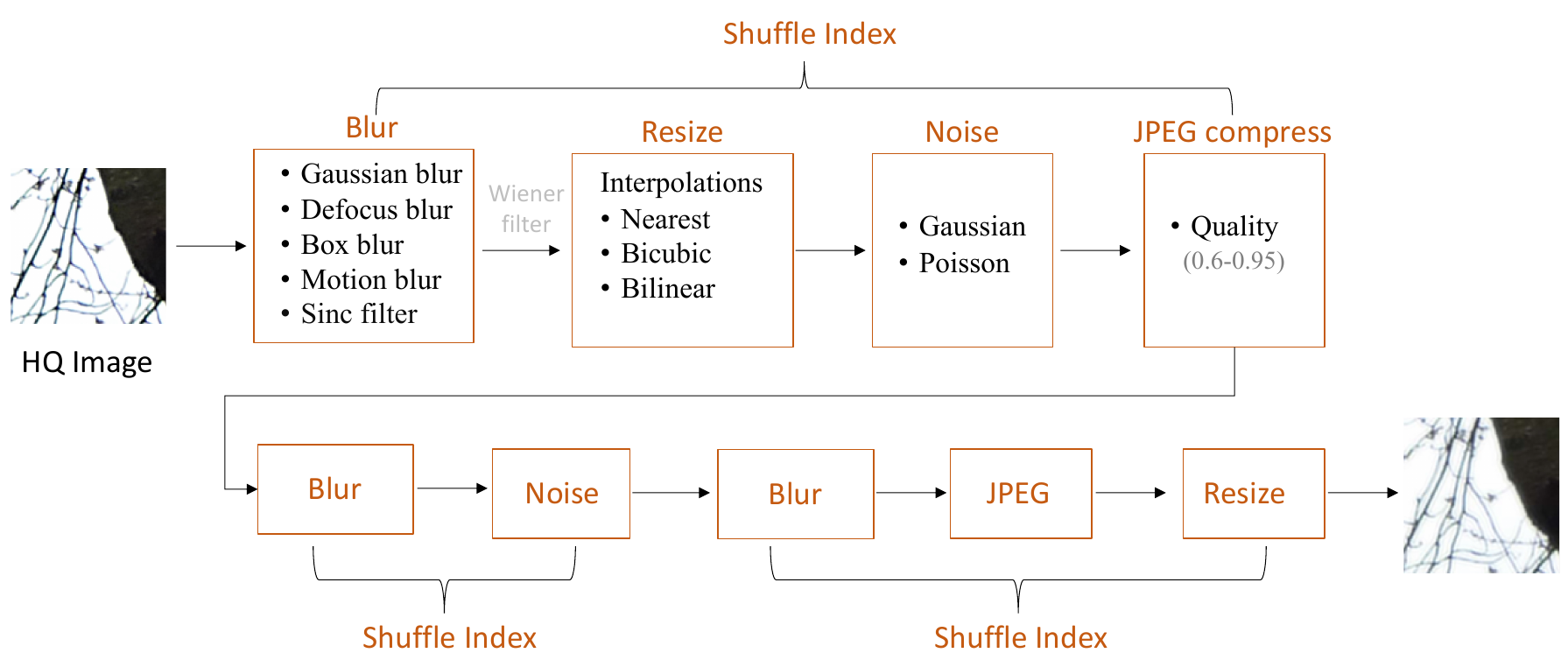}
\end{center}
    \caption{Overview of the synthetic low-quality image generation proposed by team DACLIP-IR.}
\label{fig:deg-pipeline}
\end{figure}

The model is built upon the IR-SDE~\cite{luo2023image} and DACLIP-UIR~\cite{luo2023controlling}. Since no training datasets are provided in this challenge, the team chooses to generate LQ images using a similar pipeline as in Real-ESRGAN~\cite{wang2021real} but with an index-shuffling strategy, as shown in Figure~\ref{fig:deg-pipeline}. Based on the synthetic dataset, they retrain DA-CLIP to enhance LQ features by minimizing an $\ell_1$ distance between LQ embeddings and HQ embeddings. Then they incorporate the enhanced LQ embeddings into IR-SDE with cross-attention to restore HQ images, similar to DA-CLIP~\cite{luo2023controlling}. In addition, they propose a posterior sampling approach for IR-SDE that improves both fidelity and perceptual performance. To further improve the generalization ability, they first train the model on the LSDIR dataset~\cite{li2023lsdir} and then finetune it on a mixed dataset with both synthetic and real-world images for phase two and phase three. Note that they use the same model for phase two and phase three, but take the original reverse-time SDE for phase three for better visual performance (small noise makes the photo look more realistic).

\textbf{Specific training details for phase two:} The team adds the paired validation dataset in phase one to further fine-tune the model, which improves a lot across all metrics.

\textbf{Specific training details for phase three:} They use the same model trained from phase two for phase three. To make the image look non-smooth and oil-painted, they use the original reverse-time SDE during inference.

\subsection{Team TongJi-IPOE}

Team TongJi-IPOE proposed a DRBFormer-StableSR fusion Network for restoring any image model in the Wild.

\begin{figure}[t]
\begin{center}
\includegraphics[width=1.\linewidth]{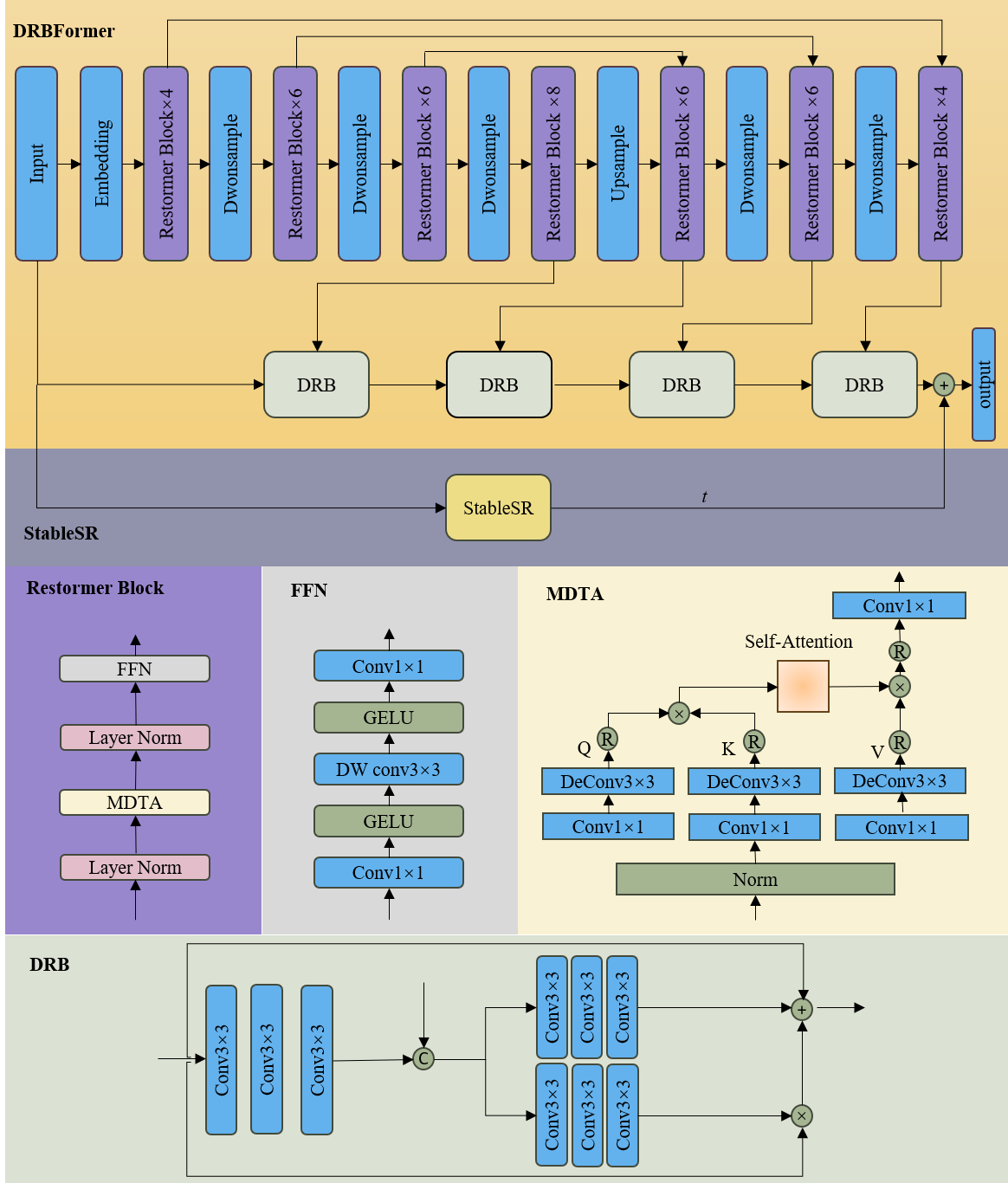}
\end{center}
    \caption{The overall architecture of the proposed method from team TongJi-IPOE.}
\label{fig:tongji}
\end{figure}

\noindent\textbf{Method.}  The overall architecture is shown in Figure \ref{fig:tongji}. The proposed network consists of two parts: DRBFormer image restoration network and StableSR \cite{Wang2023ExploitingDP} image SR network. DRBFormer uses Restormer Blocks as the backbone. Inspired by \cite{Zamir2021RestormerET}, a multi-scale dynamic residual module DRB is designed in the decoding network to better to better handle the varying blur \cite{9879086}. Considering that Diffusion priors can improve the performance of restored images, the network adopts the fusion method of Eq. (\ref{eq1}) for image restoration. Due to the randomness of the diffusion model, the generated image may deviate from the real situation, so the adjustable coefficient $t$ was set to 0.9 in this competition.\\
\begin{align}\label{eq1}
\hat{I}  & = t*DRBFormer(I_{blur})+(1-t)*StableSR(I_{blur})
\end{align}
where $\hat{I}$ is the result of restoration, $t\in \left [0,1\right]$ is adjustable coefficient and $I_{blur}$ is blurryimage.

\noindent\textbf{Training strategy.} In total, four datasets are used including DPDD\cite{Abuolaim2020DefocusDU}, SIDD\cite{8578280}, GoPro\cite{Nah2016DeepMC} and NH-HAZE\cite{9150807}. To train the models with images, the training dataset is augmented with random clipping. The details of the training steps are as follows:\\
\indent1. Pretraining on combined datasets. Ground truth patches of size 128$\times$128 are randomly cropped from Ground truth images, and the mini-batch size is set to 8. The model is trained by minimizing weighted L1 loss and perceptual loss function with Adam optimizer. The initial learning rate is set to 3$\times$$10^-4$ and the total number of iterations is 392k.\\
\indent2. Finetuning on combined datasets. For the model to adapt to higher resolution image processing, crop the image to 160$\times$160,192$\times$192,256$\times$256,320$\times$320,384$\times$384, and set the mini-batch size to [5,3,2,1,1]. The model is trained by minimizing weighted L1 loss and perceptual loss function with Adam optimizer. The initial learning rate is set to 3$\times$$10^-4$ and adjusted by cosine annealing.  The total number of iterations is 208k. 

\subsection{Team ImagePhoneix}

\begin{figure}[ht]
\centering
\includegraphics[scale=0.25]{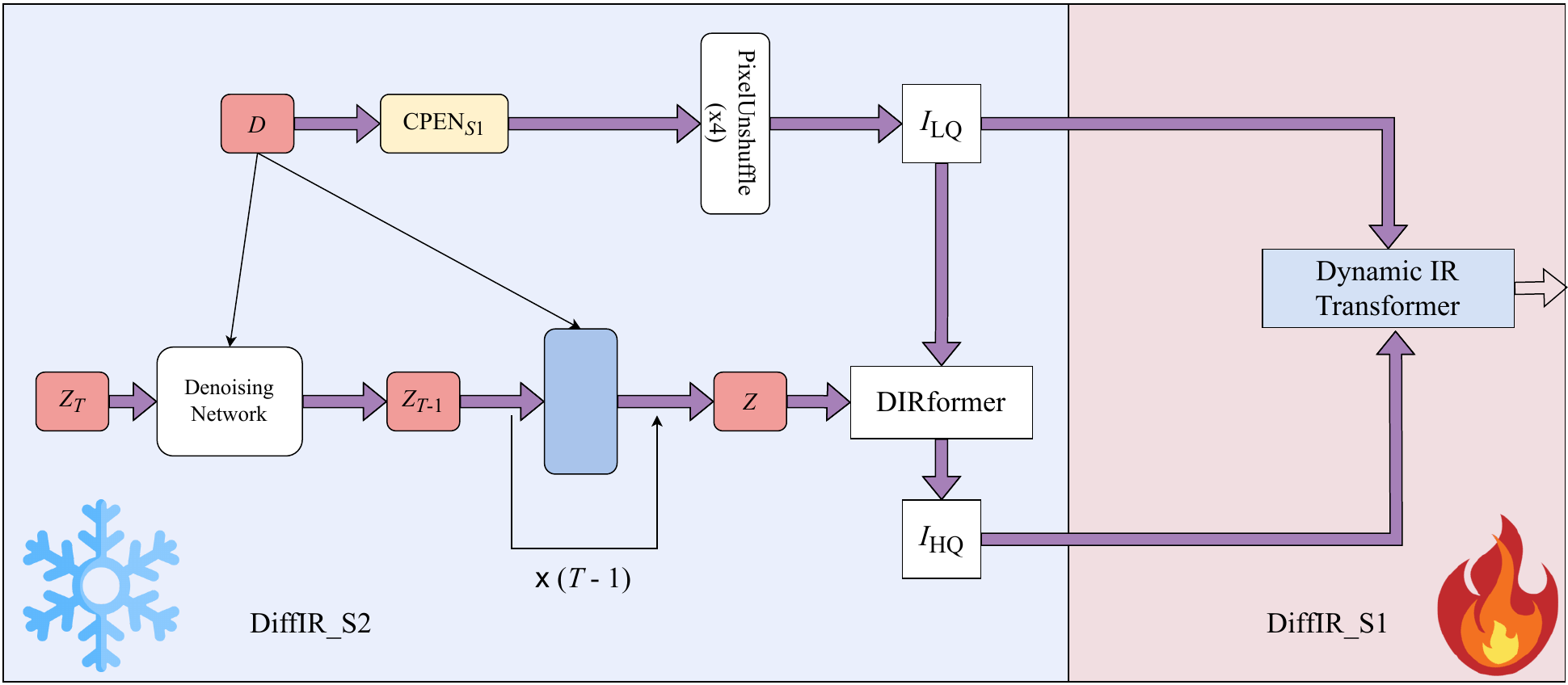} 
\caption{The technical pipeline adopted by the team ImagePhoneix.}
\label{Fig.1}
\vspace{-2mm}
\end{figure}

Team ImagePhoneix adopted DiffIR \cite{xia2023diffir} as the baseline network, as shown in Figure \textcolor{red}{1}. They froze the ``stage 2" of the DiffIR and fine-tune its ``stage 1'' network on the provided LR-HR image pairs.

\textbf{Implementation details}. 

With provided image pairs, they first cropped them into sub-images of the size $400\times 400$ for accelerating I/O speed, resulting in a total number of $2500$ sub-images. To fine-tune the pre-trained model, all the sub-images are cropped into image patches with the size $256\times 256$. They randomly flipped and rotated the input images for data augmentation. Adam algorithm is adopted with $\beta_{1}=0.9$ and $\beta_{2}=0.99$ to update the model parameters. They set the initial learning rate and the total number of iterations to $1\times 10^{-4}$ and $1\times 10^{5}$, respectively. However, the encoder is updated with a different strategy, which updates the model parameters of the encoder in $2.5\times 10^{4}$ iterations and sets the initial learning rate to $2\times 10^{-4}$. The learning rate of the encoder is decay with a factor of $0.1$ in the $1.5\times 10^{4}$-th iteration. Different from the encoder, the learning rate of the image generator is decay with a factor of $0.5$ at the $8.0\times 10^{4}$-th iteration.

In Phase II, the evaluation metric is a linear combination of the reconstruction and perceptual measurements. To handle this issue, The team adopted a hybrid loss function to fine-tune the model, which involves $L_{1}$ loss, perceptual loss based on VGG features $L_{\text{vgg}}$, adversarial loss $L_{\text{GAN}}$, and Kullback–Leibler divergence $L_{\text{KL}}$. The total loss is defined as $\mathcal{L}=\lambda_{1}L_{1} + \lambda_{2}L_{\text{vgg}} + \lambda_{3}L_{\text{GAN}} + \lambda_{4}L_{\text{KL}}$, where $L_{1}$ loss measure the reconstruction error of the generated images, $L_{\text{vgg}}$ aims to improve the perceptual quality of images, $L_{\text{GAN}}$ and $L_{\text{KL}}$ measure the distribution distance between the generated images and the ground-truth images in the spatial and latent spaces, respectively. $\lambda_{1}, \lambda_{2}, \lambda_{3}$, and $\lambda_{4}$ are hyper-parameters to balance the distortion and perceptual quality of images and set to $1.0$ in this Phase.

\subsubsection{Phase III: Evaluation on Subjective Measurements}

In Phase III, the team aims to improve the perceptual quality of generated images. Instead of using perceptual loss based on the VGG features, they adopt the robust distribution loss \cite{ni2024misalignment} which minimizes the distribution distance between the generated images and the ground-truth images based on Fast Fourier transform (FFT). Given the generated image $x$ and the ground-truth image $y$, the robust distribution loss $L_{\text{freq}}$ is defined as follows:
\begin{equation}
    L_{\text{freq}}(x, y) = L_{\text{WD}}(\mathcal{A}_{x}, \mathcal{A}_{y})+ \lambda_{\text{phase}}L_{\text{WD}}(\mathcal{P}_{x}, \mathcal{P}_{y}),
\end{equation}
where $\mathcal{A}_{x}=\vert\mathcal{F}(x)\vert$ and $\mathcal{A}_{y}=\vert\mathcal{F}(y)\vert$ denote the frequency spectrum of the images $x$ and $y$ via FFT $\mathcal{F}$, respectively. $\mathcal{P}_{x}$ and $\mathcal{P}_{y}$ represent the phase of $\mathcal{F}(x)$ and $\mathcal{F}(y)$, respectively. $L_{\text{WD}}$ is the Wasserstein distance, and $\lambda_{\text{phase}}$ is the hyper-parameter that is set to $0.1$ in the fine-tuning procedure.

\subsection{Team HIT-IIL}

The team HIT-IIL used the degradation process of Real-ESRGAN~\cite{wang2021real} and replaced the backbone with Restormer~\cite{Restormer}. For phase 2, they only trained a Real-ESRGAN x1plus model with an additional lpips loss. For phase 3, they used the backbone of Restormer to train a new x1model and averaged the results with weights 0.8 and 0.2, respectively.

They use DF2K (DIV2K and Flickr2K) datasets to train the model. For pre-processing, they use a multi-scale strategy, i.e., they downsample HR images to obtain several Ground-Truth images with different scales. They then crop DF2K images into sub-images for faster IO and processing.

\subsection{Team MARSHAL}

\subsubsection{Methods details}		
The team observed that the input images and evaluation criteria of the two phases are different. The input images in phase 2 have higher quality. The evaluation criteria for this phase are based on reference evaluation indicators. The input quality of phase 3 is relatively low, with a more serious blur. This phase uses the method of manual scoring to select images with better visual effects as the winners. Taking into account the existing solutions, the team decided to adopt a gan-based approach in phase 2 to obtain higher objective indicator scores. In phase 3, a diffusion-based approach is adopted to make the results more visually appealing.

\subsubsection{Phase 2}
The organizers provided 100 pairs of training images whose input quality and imaging style are similar to the test set of the first stage. Therefore, they chose DiffIR \cite{xia2023diffir} for this stage. It only uses the diffusion process to model the condition branch, and the main network is trained using the GAN loss, so it rarely destroys local details (such as text, small faces), and can obtain a higher objective evaluation index. They directly use the pre-trained model of DiffIR and fine-tune it with the paired dataset provided by the organizer, so that the team can quickly obtain a good result. The whole finetuning process continues 6.6 k iterations with a batch size of 48. In addition, in the process of preparing the dataset, they adopted a multi-scale downsampling strategy, hoping that the model could gain knowledge of different scales. The downsampling scales are set to 0.75, 0.5, and 0.33, respectively.

\subsubsection{Phase 3}
\begin{figure}[t]
  \centering
  \includegraphics[scale=0.5]{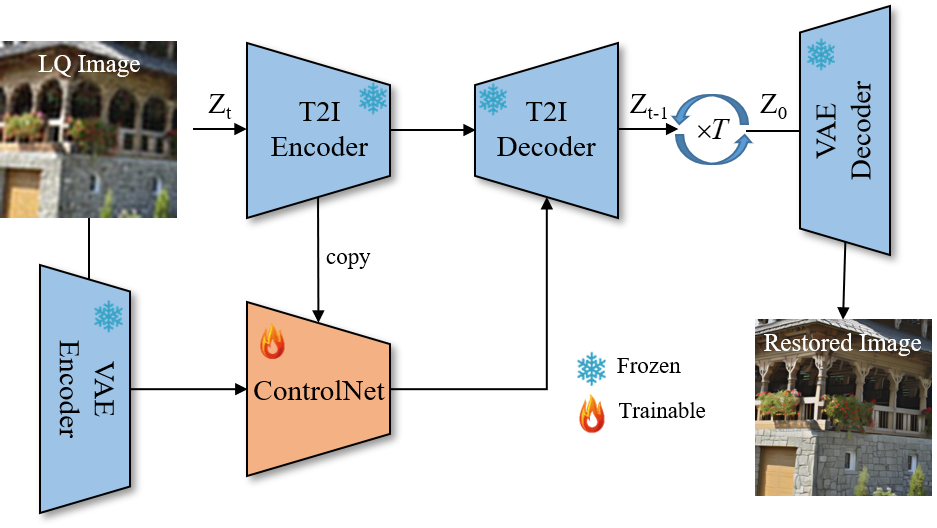}
  \caption{The pipeline of the solution proposed by team MARSHAL.}
  \label{fig:fangan}
\end{figure}

\begin{figure}[t]
  \centering
  \includegraphics[scale=0.3]{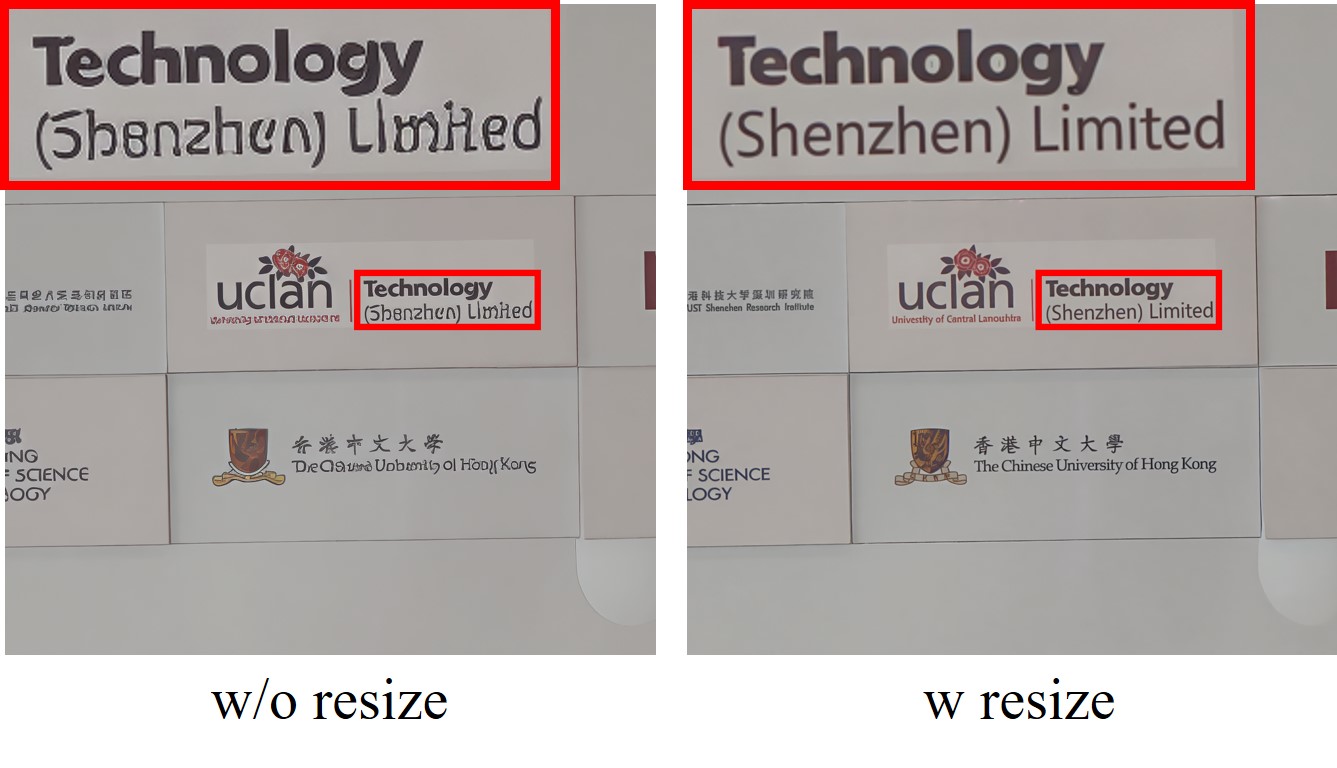}
  \caption{Comparison of resize strategies on small text scenarios in the solution proposed by team MARSHAL.}
  \label{fig:resize}
\end{figure}

In phase 3, the test set provided by the organizer has a significant domain gap from the test set of phase 2, and the degradation is more severe. The team thinks they cannot directly use the model that performs well in phase 2 to obtain good visual results in the second stage, so they switched to using the methods \cite{wang2023exploiting, yang2023pixel, wu2023seesr, sun2023improving} of the pre-training diffusion model. As shown in Fig. \ref{fig:fangan}, the team chooses the popular ControlNet \cite{zhang2023adding} as the solution. Following \cite{lin2023diffbir}, they use pretrained VAE encoder as the image encoder. In terms of training data, they choose LSDIR \cite{li2023lsdir}, which contains tens of thousands of texture-rich images. As for data degradation, to match the more severe degradation of the test set, they choose realesrgan's \cite{wang2021real} degradation pipeline to synthesize paired data. They train the model with a batch size of 32 for 100k iterations. In the inference stage, the team resizes the input to 2048 before feeding it into the model, which aims to preserve small structures like texts as shown in Fig. \ref{fig:resize}. The team also adopts the LRE strategy proposed in \cite{wu2023seesr} to improve fidelity. The pre-trained diffusion model in this solution is SD2-base \cite{rombach2022high}.

\section{Acknowledgments}

This work was partially supported by the Humboldt Foundation. We thank the NTIRE 2024 sponsors: Meta Reality Labs, OPPO, KuaiShou, Huawei and University of W\"urzburg (Computer Vision Lab).

\section{Appendix: Teams and affiliations}
\label{appendix}

\textbf{NTIRE 2024 Team}
\flushleft

\noindent\textit{\textbf{Challenge:}} 

\noindent NTIRE 2024 Restore Any Image Model (RAIM) in the Wild

\noindent\textit{\textbf{Organizers:}}

\noindent Jie Liang$^1$ (liang27jie@gmail.com)

\noindent Qiaosi Yi$^{1,2}$ (qiaosi.yi@connect.polyu.hk)

\noindent Shuaizheng Liu$^{1,2}$ (shuaizhengliu21@gmail.com)

\noindent Lingchen Sun$^{1,2}$ (ling-chen.sun@connect.polyu.hk)

\noindent Xindong Zhang$^{1}$ (17901410r@connect.polyu.hk)

\noindent Hui Zeng$^{1}$ (cshzeng@gmail.com)

\noindent Prof. Lei Zhang$^{1,2}$ (cslzhang@comp.polyu.edu.hk)

\noindent Prof. Radu Timofte$^3$ (radu.timofte@uni-wuerzburg.de)

\noindent\textit{\textbf{Affiliations:}}

\noindent $^1$ OPPO Research Institute

\noindent $^2$ The Hong Kong Polytechnic University 

\noindent $^3$ Computer Vision Lab, University of W\"urzburg, Germany

~\\

\textbf{Team MiAlgo}

\noindent\textit{\textbf{Members:}}

Yibin Huang (huangyibin@xiaomi.com)

Shuai Liu, Yongqiang Li, Chaoyu Feng, Xiaotao Wang, Lei Lei

\noindent\textit{\textbf{Affiliations:}}

Xiaomi Inc., China

~\\

\textbf{Team Xhs-IAG}

\noindent\textit{\textbf{Members:}}

Yuxiang Chen\textsuperscript{\rm 1} (chenyuxiang@xiaohongshu.com)

Xiangyu Chen\textsuperscript{\rm 2,3}, Qiubo Chen\textsuperscript{\rm 1}

\noindent\textit{\textbf{Affiliations:}}

\textsuperscript{\rm 1} Xiaohongshu, 

\textsuperscript{\rm 2} University of Macau, 

\textsuperscript{\rm 3} Shenzhen Institutes of Advanced Technology, Chinese Academy of Sciences

~\\

\textbf{Team So Elegant}

\noindent\textit{\textbf{Members:}}

Jiaxu Chen

Fengyu Sun (sunfengyu@s.upc.edu.cn), Mengying Cui

\noindent\textit{\textbf{Affiliations:}}

China University of Petroleum (East China)

~\\

\textbf{Team IIP\_IR}

\noindent\textit{\textbf{Members:}}

Zhenyu Hu (zhenyuhu@whu.edu.cn), \\ 
Jingyun Liu, Wenzhuo Ma, Ce Wang, Hanyou Zheng, Wanjie Sun, Zhenzhong Chen

\noindent\textit{\textbf{Affiliations:}}

School of Remote Sensing and Information Engineering, Wuhan University

~\\

\textbf{Team DACLIP-IR}

\noindent\textit{\textbf{Members:}}

Ziwei Luo (ziwei.luo@it.uu.se)

Fredrik K. Gustafsson, Zheng Zhao, Jens Sjölund, Thomas B. Schön

\noindent\textit{\textbf{Affiliations:}}

Department of Information Technology, Uppsala University

~\\

\textbf{Team TongJi-IPOE}

\noindent\textit{\textbf{Members:}}

Xiong Dun 

Pengzhou Ji (jipengzhoudrew@163.com), Yujie Xing, Xuquan Wang, Zhanshan Wang, Xinbin Cheng

\noindent\textit{\textbf{Affiliations:}}

Institute of Precision Optical Engineering, School of Physics Science and Engineering, Tongji University

~\\

\textbf{Team ImagePhoneix}

\noindent\textit{\textbf{Members:}}

Jun Xiao$^{1}$ (jun.xiao@connect.polyu.hk)

Chenhang He$^{1}$, Xiuyuan Wang$^{1}$, Zhi-Song Liu$^{2}$

\noindent\textit{\textbf{Affiliations:}}

$^{1}$ The Hong Kong Polytechnic University

$^{2}$ Lappeenranta-Lahti University of Technology

~\\

\textbf{Team HIT-IIL}

\noindent\textit{\textbf{Members:}}

Zimeng Miao (2214177602@qq.com)

Zhicun Yin, Ming Liu, Wangmeng Zuo

\noindent\textit{\textbf{Affiliations:}}

School of Computer Science and Technology, Harbin Institute of Technology

~\\

\textbf{Team MARSHAL}

\noindent\textit{\textbf{Members:}}

Rongyuan Wu (rong-yuan.wu@connect.polyu.hk)

Shuai Li

\noindent\textit{\textbf{Affiliations:}}

The Hong Kong Polytechnic University
{
    \small
    \bibliographystyle{ieeenat_fullname}
    \bibliography{main}
}


\end{document}